\documentclass[lettersize,journal]{IEEEtran}
\usepackage{cite}
\usepackage{amsmath,amssymb,amsfonts}
\usepackage{algorithmic}
\usepackage{graphicx}
\usepackage{textcomp}
\usepackage[center]{caption}
\usepackage{booktabs}
\usepackage{mathtools}
\usepackage{listings}
\usepackage{setspace}
\usepackage{lipsum,multicol}
\usepackage{adjustbox}
\usepackage{url}
\usepackage{bbding}
\usepackage[subrefformat=parens,labelformat=parens]{subfig}
\usepackage{refstyle}
\usepackage{floatrow}
\usepackage{cuted}
\usepackage{array}
\usepackage{textcomp}
\usepackage{stfloats}
\usepackage{url}
\usepackage{verbatim}
\def\BibTeX{{\rm B\kern-.05em{\sc i\kern-.025em b}\kern-.08em
    T\kern-.1667em\lower.7ex\hbox{E}\kern-.125emX}}
\usepackage{balance}
\newcolumntype{C}{>{\centering\arraybackslash}X} 
\usepackage{multirow}
\usepackage{hyperref}       

\begin{document}
%
\title{The Effects of Partitioning Strategies on Energy Consumption in Distributed CNN Inference at The Edge}

%

\author{ 

\begin{tabular}{ccc}
\href{https://orcid.org/0000-0002-7383-321X}{\includegraphics[scale=0.06]{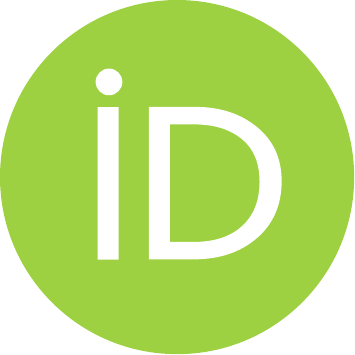}\hspace{1mm}Erqian Tang} & \href{https://orcid.org/0000-0003-4540-9013}{\includegraphics[scale=0.06]{orcid.pdf}\hspace{1mm}Xiaotian Guo} & \href{https://orcid.org/0000-0001-6006-9366}{\includegraphics[scale=0.06]{orcid.pdf}\hspace{1mm}Todor Stefanov} \\
 Leiden University          & University of Amsterdam, Leiden University             & Leiden University         \\

e.tang@liacs.leidenuniv.nl       & x.guo3@uva.nl          & t.p.stefanov@liacs.leidenuniv.nl         
\end{tabular}
}


\maketitle

 \begin{abstract}
Nowadays, many AI applications utilizing resource-constrained edge devices (e.g., small mobile robots, tiny IoT devices, etc.) require Convolutional Neural Network (CNN) inference on a distributed system at the edge due to limited resources of a single edge device to accommodate and execute a large CNN. There are four main partitioning strategies that can be utilized to partition a large CNN model and perform distributed CNN inference on multiple devices at the edge. However, to the best of our knowledge, no research has been conducted to investigate how these four partitioning strategies affect the energy consumption per edge device.  Such an investigation is important because it will reveal the potential of these partitioning strategies to be used effectively for reduction of the per-device energy consumption when a large CNN model is deployed for distributed inference at the edge. Therefore, in this paper, we investigate and compare the per-device energy consumption of CNN model inference at the edge on a distributed system when the four partitioning strategies are utilized. The goal of our investigation and comparison is to find out which partitioning strategies (and under what conditions) have the highest potential to decrease the energy consumption per edge device when CNN inference is performed at the edge on a distributed system. 
\end{abstract}


%
\IEEEpeerreviewmaketitle

\section{Introduction}
\label{sec:Intro}

Convolutional Neural Networks (CNNs) have been intensively researched and widely used in application areas such as image recognition and natural language processing due to their ability to process data in large scale and with high classification accuracy after training~\cite{b1}. A trained CNN model can be deployed on different kinds of hardware platforms to perform CNN inference on actual data within different applications. The CNN inference is usually computation-intensive and some CNN models are huge, i.e., they require a lot of memory and consume a lot of energy to perform the inference. Powerful hardware like a server or a desktop computer usually has enough memory and sufficient power/energy supply to perform the inference of large CNN models. However, some applications need to perform CNN inference on resource-constrained edge devices that may have very limited energy budget to execute a CNN model for a certain period of time. For example: (1) In an Internet of Things (IoT) smart home application, edge devices like a light controller or a temperature conditioner need to deploy and execute a natural language processing model in order to recognize human input voices and generate output commands accordingly; (2) In a military mobile robots application, some tiny size robots like robotic ants and robotic insects need to invade specific places under surveillance to detect illegal objects (e.g., weapons) and terrorist activities, utilizing image recognition models. In such application scenarios, the IoT devices and tiny robots are usually powered by very small batteries in order to obtain portability, and thus deploying a CNN model and performing the inference on a single edge device is not possible because the hardware platform has very limited resources in terms of energy budget to execute the entire CNN model for a relatively long period of time, e.g., the time needed to complete the robot mission. One approach to solve this issue is to perform CNN model compression~\cite{b2, b3, b4} (e.g., pruning, quantization, or knowledge distillation) but such an approach sacrifices the accuracy of the model to some extent, especially when high model compression rates are required. Another approach is to deploy only part of the CNN model on the edge device and the rest of the model on the cloud or remote server/desktop~\cite{b5, b6} but such an approach may have data privacy issues like leakage of personal information or sensitive military information. In addition, sending data from the edge device to the cloud or remote server/desktop and returning back results may lead to unacceptable CNN inference latency due to communication network delays and transmission distances. A third approach, which solves the aforementioned issues of the other two approaches, is to deploy the CNN model on a (fully) distributed system at the edge and take advantage of all available edge devices to cooperatively perform the CNN inference. The existing methodologies that use this approach to deploy a CNN model and perform inference at the edge on a distributed system can be divided in four main categories depending on the CNN model partitioning strategy which is utilized: 

\begin{enumerate}
	\item Methodologies utilizing the Data Partitioning Strategy, e.g.,~\cite{b7};
	\item Methodologies utilizing the Sequential Partitioning Strategy, e.g.,~\cite{b8, b9};
	\item Methodologies utilizing the Horizontal Partitioning Strategy, e.g.,~\cite{b10, b11, b12};
	\item Methodologies utilizing the Vertical Partitioning Strategy, e.g.,~\cite{b19}.
\end{enumerate}

Substantial amount of research has been conducted to investigate the overall system performance and memory requirements per edge device when the aforementioned four partitioning strategies are utilized to deploy a CNN model and perform inference at the edge on a distributed system. However, to the best of our knowledge, no research has been conducted to investigate how the aforementioned four partitioning strategies affect the energy consumption per edge device. Such an investigation is important because it will reveal the potential of these partitioning  strategies to be used effectively for reduction of the energy consumption per edge device when a CNN model is deployed for distributed inference at the edge. Since the energy budget of edge devices is limited, it is important for designers and developers to understand the advantages and disadvantages of the four partitioning strategies in terms of energy consumption per edge device. Therefore, in this paper, we investigate and compare the per-device energy consumption of CNN model inference at the edge on a distributed system when the four partitioning strategies are utilized. The goal of our investigation and comparison is to find out which partitioning strategies (and under what conditions) have the highest potential to decrease the energy consumption per edge device when CNN inference is performed at the edge on a distributed system. 

\subsection*{Paper contributions}
\label{sec:contrib}
In this paper, we investigate, compare, report, and discuss the energy consumption per edge device of CNN model inference at the edge when the aforementioned partitioning strategies are utilized. In order to achieve this, our main novel contributions are:
\begin{itemize}
\item We analyze the characteristics of each partitioning strategy in terms of computation/communication workload of every partition as well as communication and synchronization complexity between different partitions. Based on this analysis, we propose novel and very accurate analytical models to estimate the energy consumption of every partition when utilizing each of the four partitioning strategies for CNN inference on a distributed system at the edge; 
\item We confirm the high accuracy of our analytical models by comparing the energy consumption numbers provided by our models with measured numbers obtained by deploying a few real-world CNNs on a distributed system containing several NVIDIA Jetson TX2 embedded platforms; 
\item We utilize our accurate energy consumption analytical models in a large-scale experiment including nine representative real-world CNN models from the ONNX model zoo library~\cite{b14} to investigate which partitioning strategy has the highest potential to decrease the per-device energy consumption for every one of the nine CNN models inferred on distributed devices at the edge. 
\end{itemize}
 
The remainder of the paper is organized as follows: Section~\ref{sec:related_work} describes the related work. Section~\ref{sec:Background} provides some background information needed to understand the contributions of this paper. Section~\ref{sec:model} presents in detail our analytical models that are used in our investigation of the energy consumption of CNN model inference at the edge on a distributed system. Section~\ref{sec:validation} validates the accuracy of our energy consumption analytical models. Section~\ref{sec:investigation} presents and discusses our large-scale investigation and obtained results. Section \ref{sec:Conclusions} ends the paper with conclusions.

\section{Related Work}
\label{sec:related_work}

To the best of our knowledge, our work is the first trying to investigate how the four existing partitioning strategies affect the energy consumption per edge device when a CNN model is deployed for distributed inference at the edge. As mentioned earlier, such an investigation is important because it will reveal the potential of these partitioning strategies to be used effectively for reduction of the per-device energy consumption. Therefore, in this section, we discuss some related works dealing with CNN inference on distributed system and the energy consumption of CNN inference on Multi-processor System-on-Chip (MPSoC).

The research works presented in~\cite{b15, b7, b17, b18, b19} investigate the system performance and memory consumption of CNN inference on a distributed system at the edge. The contributions of these works are to design a new or evaluate existing mapping/partitioning strategies for improving the efficiency in terms of system performance and memory consumption when a CNN is deployed for inference on a distributed system at the edge. However, these  works do not consider the per-device energy consumption in such distributed CNN inference. In a distributed system at the edge, the available energy budget for edge devices is typically very limited and it is important for such edge devices to meet this limited energy budget. In contrast to~\cite{b15, b7, b17, b18, b19}, our work reveals the potential of the existing partitioning strategies to be used effectively for reduction of the energy consumption per edge device when a CNN model is deployed for distributed inference at the edge.

The works in~\cite{b13, b20, b21, b22} investigate the energy consumption of CNN inference on MPSoC. The contribution of this body of works is to design a new or evaluate existing mapping/partitioning strategies for improving the efficiency in terms of energy consumption when CNN inference is deployed on a single MPSoC. In order to do that, real system implementations are realized and analytical models are devised for different CNN applications mapped on a single embedded hardware platform such as NVIDIA Jetson TX2 or Raspberry Pi. However, these research works do not consider per-device energy consumption of CNN inference on a distributed system which consists of multiple edge devices. In such distributed system at the edge, data communication between edge devices will consume much more energy compared to data communication within a single MPSoC. In contrast to~\cite{b13, b20, b21, b22}, our work reveals the potential of the existing partitioning strategies to be used effectively for reduction of the energy consumption per edge device when a CNN model is deployed for distributed inference on multiple edge devices. Moreover, our analytical energy consumption models consider both the energy needed for computation per edge device and the energy needed for data communication between different edge devices. Our real system implementations on hardware platforms validate our energy models' accuracy and our models calibration by profiling ensures high accuracy of the obtained results during the large-scale investigation, we present in this paper.

\section{Background}
\label{sec:Background}
In this section, first we briefly introduce the CNN computational model and related notations used throughout the paper. Next, the four existing CNN model partitioning strategies, investigated in this paper, are described. Finally,  our experimental setup is presented including the real-world CNN models used in our investigation and the distributed hardware platform utilized for validation and calibration of our analytical energy consumption models.

\subsection{CNN computational model}
\label{CNN_model}
A Convolutional Neural Network (CNN) is a computational model~\cite{b23}, commonly represented as a directed acyclic computational graph $CNN\left ( L,E \right )$ with a set of nodes $L$, also called layers, and a set of edges $E$. Each layer $l_i \in L $ accepts input data $X_i$ and provides output data $Y_i$. To obtain the output data $Y_i$ from the input data $X_i$, layer $l_i$ moves along $X_i$ with sliding window $\Theta _i$ and stride $s_i$, applying operator $op_i$ to an area of $X_i$, covered by $\Theta _i$. The areas covered by $\Theta _i$ can overlap. If input data $X_i$ cannot be covered by sliding window $\Theta _i$ integer number of times, layer $l_i$ crops or extends $X_i$ with padding $pad_i$~\cite{b1} and processes cropped/extended input data $X'_i$, which can be covered by sliding window $\Theta _i$ integer number of times.

The layers input and output data is stored in multidimensional arrays, called tensors. In this paper, each input/output tensor $T$ has the
format $T \left [ H^T ,W^T ,C^T \right ]$, where $H^T$ is the tensor height, $W^T$ is the tensor width, $C^T$is the number of channels. We define a layer as a tuple $l_i = (X_i, Y_i, \Theta _i, op_i, s_i, pad_i)$, where

\begin{itemize}
	\item $X_i^{\left [ H^X_i, W^X_i, C^X_i \right ]}$ is the input data tensor of $l_i$;
	\item $Y_i^{\left [ H^Y_i, W^Y_i, C^Y_i \right ]}$ is the output data tensor of $l_i$;
	\item $\Theta _i^{\left [ N^\Theta_i, H^\Theta_i, W^\Theta_i, C^\Theta_i \right ]}$ is the sliding window of $l_i$, $N^\Theta_i = C^Y_i$ is the number of neurons in $l_i$, and $C_i^\Theta = C_i^X$;
	\item $s_i$ is the stride, with which $l_i$ moves over $X_i$;
	\item $op_i$ is the operator of $l_i$;
	\item $pad_i$ is the padding of $l_i$;
\end{itemize}

\subsection{CNN Model Partitioning Strategies}
\label{partitioning_strategies}

\begin{figure*}[!t]
	\begin{minipage}{0.49\columnwidth}%
		\subfloat[Data Partitioning Strategy]{
			\label{fig:data_partition}
			\includegraphics[width=\linewidth]{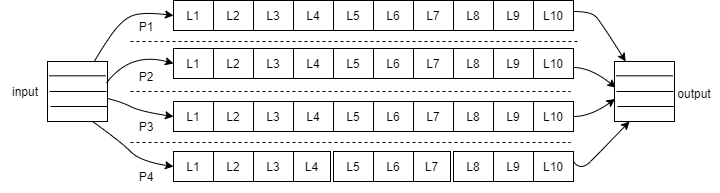}
		}%
	\end{minipage}
	\begin{minipage}{0.49\columnwidth}%
		\subfloat[Sequencial Partitioning Strategy]{
			\label{fig:sequantial_partition}
			\includegraphics[width=\linewidth]{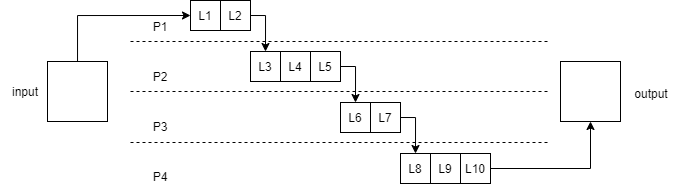}
		}%
	\end{minipage} 
	\begin{minipage}{0.49\columnwidth}%
	\vspace{+0.35cm}
		\subfloat[Horizental Partitioning Strategy]{
			\label{fig:horizental_partition}
			\includegraphics[width=\linewidth]{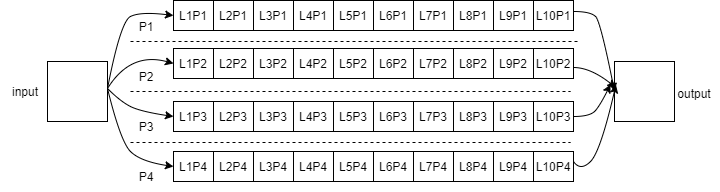}
		}%
	\end{minipage} 
	\begin{minipage}{0.49\columnwidth}%
		\subfloat[Vertical Partitioning Strategy]{
			\label{fig:vertical_partition}
			\includegraphics[width=\linewidth]{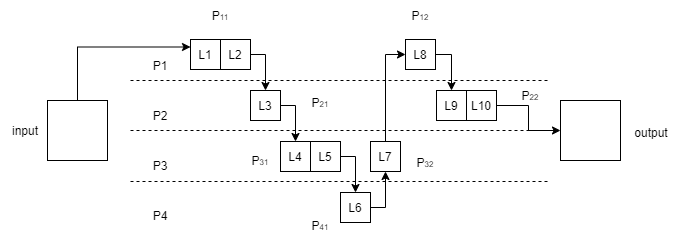}
		}%
	\end{minipage}
	\caption{Fully distributed CNN model partitioning strategies}
	\label{fig:methodologies}
\end{figure*}

{\bf Data Partitioning Strategy:} The main features of the data partitioning strategy are shown in Row 3 of Table \ref{tab:partitioning_methods}. This strategy may partition the input data given to every CNN layer $l_i$ whereas the weights of every CNN layer are not partitioned. Each CNN model partition $p_j$ includes all layers of the CNN model, where the layers operate on only part of their input data because the input data to a layer is partitioned. Different CNN model partitions have dependencies because some parts of the input data to layer $l_i$ of partition $p_j$ may have to be shared with layers from other partition. Figure~\subref*{fig:data_partition} shows a CNN model with ten layers $l_1$ to $l_{10}$ that is partitioned into four partitions $p_1$ to $p_4$. The input image is partitioned into four parts by a master node and each input image part is forwarded to one partition for processing. An example of a methodology utilizing this partitioning strategy to deploy a CNN model and perform inference on a distributed system at the edge can be found in~\cite{b7}.

{\bf Sequential Partitioning Strategy:} The main features of the sequential partitioning strategy are shown in Row 4 of Table~\ref{tab:partitioning_methods}. This strategy partitions the layers of a CNN model such that each partition $p_j$ includes consecutive CNN layers. The input data given to a CNN layer $l_i$ and its weights are not partitioned. For example, Figure~\subref*{fig:sequantial_partition} shows a CNN model with ten layers $l_1$ to $l_{10}$ that is partitioned into four partitions $p_1$ to $p_4$. An example of methodologies utilizing this partitioning strategy to deploy a CNN model and supporting inference on a distributed system at the edge can be found in~\cite{b8, b9}.

\begin{table}[!t]
	\resizebox{\columnwidth}{!}{
		\begin{tabular}{|c|c|c|c|c|}
			\hline
			\multirow{2}{*}{CNN model partitioning strategy} & \multicolumn{2}{c|}{items of a layer to partition} & \multicolumn{2}{c|}{layers to include per partition} \\ \cline{2-5} 
			& input data                & weights                & all                   & consecutive                  \\ \hline
			Data Partitioning Strategy                       & Yes                       & No                     & Yes                   & Yes                          \\ \hline
			Sequentially Partitioning Strategy               & No                        & No                     & No                    & Yes                          \\ \hline
			Horizontal Partitioning Strategy                 & No                        & Yes                    & Yes                   & Yes                          \\ \hline
			Vertical Partitioning Strategy                   & No                        & No                     & No                    & No                           \\ \hline
	\end{tabular}}
	\caption{Features of CNN model partitioning strategies}
	\label{tab:partitioning_methods}
\end{table}
{\bf Horizontal Partitioning Strategy:} The main features of the horizontal partitioning strategy are shown in Row 5 of Table~\ref{tab:partitioning_methods}. This strategy partitions the weights of every CNN layer $l_i$ whereas the input data given to a CNN layer is not partitioned. Each CNN model partition $p_j$ includes all layers of the CNN model, where the layers operate with only part of their weights because the weights of a layer are partitioned. Therefore, every CNN layer $l_i$ is divided in several parts $l_ip_j$ and distributed onto several partitions. Different parts of the same CNN layer have to communicate and synchronize with each other because all the output data from part $l_ip_j$ has to be concatenated with the output data from the other parts of the same layer. For example, Figure~\subref*{fig:horizental_partition} shows a CNN model with ten layers $l_1$ to $l_{10}$ that is partitioned into four partitions $p_1$ to $p_4$. An example of methodologies utilizing this partitioning strategy to deploy a CNN model and perform inference on a distributed system at the edge can be found in~\cite{b10, b11, b12}.

{\bf Vertical Partitioning Strategy:} The main features of the vertical partitioning strategy are shown in Row 6 of Table~\ref{tab:partitioning_methods}. This strategy may partition the layers of a CNN model such that each partition $p_j$ includes non-consecutive CNN layers. The input data given to a CNN layer $l_i$ and its weights are not partitioned. For example, Figure~\subref*{fig:vertical_partition} shows a CNN model with ten layers $l_1$ to $l_{10}$ that is partitioned into four partitions $p_1$ to $p_4$. An example of a methodology utilizing this partitioning strategy to deploy a CNN model and perform inference on a distributed system at the edge can be found in~\cite{b19}.

\subsection{CNN models used in the investigation}
\label{sec:CNNs}
Open Neural Network Exchange (ONNX) is an open standard format for representing machine learning models. ONNX is supported by a community of partners who have implemented it in many frameworks and tools. ONNX Model Zoo~\cite{b14} is a collection of pre-trained, state-of-the-art models in the ONNX format. For our large-scale investigation, we select the nine real-world CNN models, shown in Table~\ref{tab:CNNs}, from the ONNX model zoo that take images as input for CNN inference. These CNN models are utilized in different applications and have diverse number and type of layers. Such diversity leads to a diverse scale of energy consumption when these CNN models are mapped and executed on one and the same distributed system at the edge. 

\begin{table*}[!t]
	\begin{tabular}{|l|l|c|c|r|}
		\hline
		CNN model                        & \multicolumn{1}{c|}{Application}                                                  & Number of  layers & Input data size & Weights size \\ \hline\hline
		CaffeNet                         & Image Classification                                                              & 15                & $224\times224\times3$       & 232 MB      \\ \hline
		ResNet 50 & Image Classification                                                              & 124               & $224\times224\times3$       & 98 MB       \\ \hline
		SqueezeNet                       & Image Classification                                                              & 39                & $224\times224\times3$       & 5 MB        \\ \hline
		VGG 16                           & Image Classification                                                              & 23                & $224\times224\times3$       & 528 MB      \\ \hline
		VGG 19                           & Image Classification                                                              & 26                & $224\times224\times3$       & 549 MB      \\ \hline
		AlexNet                          & Image Classification                                                              & 15                & $224\times224\times3$       & 232 MB      \\ \hline
		Tiny YOLO V2                     & \begin{tabular}[c]{@{}l@{}}Object Detection \& \\ Image Segmentation\end{tabular} & 24                & $416\times416\times3$       & 60 MB       \\ \hline
		YOLO V2                          & \begin{tabular}[c]{@{}l@{}}Object Detection \& \\ Image Segmentation\end{tabular} & 24                & $416\times416\times3$       & 204 MB      \\ \hline
		Emotion\_fer                     & \begin{tabular}[c]{@{}l@{}}Body, Face \& \\ Gesture Analysis\end{tabular}         & 19                & $64\times64\times1$         & 20 MB       \\ \hline
	\end{tabular}
\caption{CNN models selected from the ONNX model zoo \cite{b14}}
\label{tab:CNNs}
\end{table*}

As shown in Column 2 of Table~\ref{tab:CNNs}, CaffeNet, ResNet 50, SqueezeNet, VGG 16, VGG 19, and AlexNet are used for image classification. Tiny YOLO V2 and YOLO V2 are used for object detection and image segmentation. Emotion\_fer is used for body, face, and gesture analysis. The number of layers, input data size, and total weights size of these nine selected CNN models are shown in Column 3, 4, and 5, respectively. As can be seen from Table~\ref{tab:CNNs}, these CNN models are sufficiently diverse, representative, and good examples to utilize in our large-scale experiment in order to investigate the energy consumption per edge device by applying our accurate analytical energy consumption models.

\subsection{Distributed hardware platform}
\label{sec:Distributed_hardware_platform}
In order to validate and confirm the accuracy of our analytical energy consumption models, we implement two CNN modes (VGG 16 and Emotion\_fer) and execute them on a distributed system which consists of four NVIDIA Jetson TX2 embedded platforms~\cite{b26} by applying the partitioning strategies introduced in Section~\ref{partitioning_strategies}. During our large-scale investigation, first, we profile each of the nine selected CNN models, introduced in Section \ref{sec:CNNs}, on one NVIDIA Jetson TX2 platform. Then, the obtained profile data is used to calibrate our analytical energy consumption models. Finally, we use the calibrated energy models to investigate the energy consumption of every one of the nine CNN models mapped on a variety of distributed systems consisting of multiple NVIDIA Jetson TX2 platforms. 

Every NVIDIA Jetson TX2 platform features six CPUs (Quad-Core ARM and Dual-Core NVIDIA Denver 2) plus one Pascal GPU. We select and utilize multiple NVIDIA Jetson TX2 platforms as edge devices in a distributed system because Jetson TX2 is a well-known and easy-to-use embedded MPSoC platform. Moreover, we can easily and accurately acquire the needed energy consumption data of each device by setting timers within the executed code and by sampling the integrated power sensors onboard an NVIDIA Jetson TX2 platform.

For the communication between different edge devices in a distributed hardware platform, we use Message-Passing Interface (MPI) \cite{b27}. MPI is a portable standard designed for parallel computing architectures which includes point-to-point message-passing, collective communications, process creation and management, extended collective operations, etc. There are two advantages of utilizing MPI for the communication between different edge devices in our distributed hardware platform. First, MPI supports various network hardware and topologies such as Infiniband and Intel Omni Path or Dragonfly. Usually, MPI provides very good performance on scalable parallel computers with specialized inter-processor communication hardware. The second advantage of utilizing MPI is its wide portability. The interface is suitable for use by fully general Multiple Instruction Multiple Data (MIMD)~\cite{b28} programs as well as programs written in the more restricted style of Single Program Multiple Data (SPMD)~\cite{b29}. MPI programs may run on distributed-memory multiprocessors, parallel workstations, and combinations of machines connected by a communication network. So, it is easy to implement MPI on top of standard Unix inter-processor communication protocols which provides portability to edge devices in a distributed system. 

\section{Energy Consumption Models}
\label{sec:model}

The energy consumption of each device in a distributed system at the edge for CNN model inference can be broken down to three parts: (1) Energy consumed by the computation operations in the CNN layers (convolution operations, activation operations, etc.), executed on the edge device; (2) Energy consumed for data communication between the CNN layers, executed on the edge device; (3) Energy consumed for data communication between the edge device and other edge devices in the distributed system. For the different partitioning strategies, described in Section~\ref{partitioning_strategies}, each partition will be mapped and executed on one edge device in the distributed system and the energy consumption per device can be very different depending on the partitioning strategy. This is because the different partitioning strategies differ in how layers, input/output data to/from layers, and weights are partitioned, thereby causing differences in the computation and communication workload of one partition and the amount of data communicated between the partition and other partitions. Based on the CNN model description, introduced in Section \ref{CNN_model}, we propose energy consumption models for the four different partitioning strategies, briefly described in Section \ref{partitioning_strategies}.

\subsection{Energy Model for Data Partitioning Strategy}
\label{sec:model_data}
For the data partitioning strategy, the weights of every CNN layer are not partitioned whereas the input data given to every CNN layer $l_i$  is partitioned along the height $H^X_i$ of the input tensor $X_i^{\left [ H^X_i ,W^X_i ,C^X_i \right ]}$ into $M$ partitions, in such a way that the output tensor $Y_i^{\left [ H^Y_i \left(j\right) ,W^Y_i ,C^Y_i \right ]}$ from every partition $j$ of the same layer $l_i$ can be concatenated with the other output tensors of partitions into one whole piece of output tensor along height $H^Y_i = \sum_{j=1}^{M} H^Y_i \left(j\right)$. 

\begin{figure*}[!t]
	\includegraphics[width=.76\linewidth]{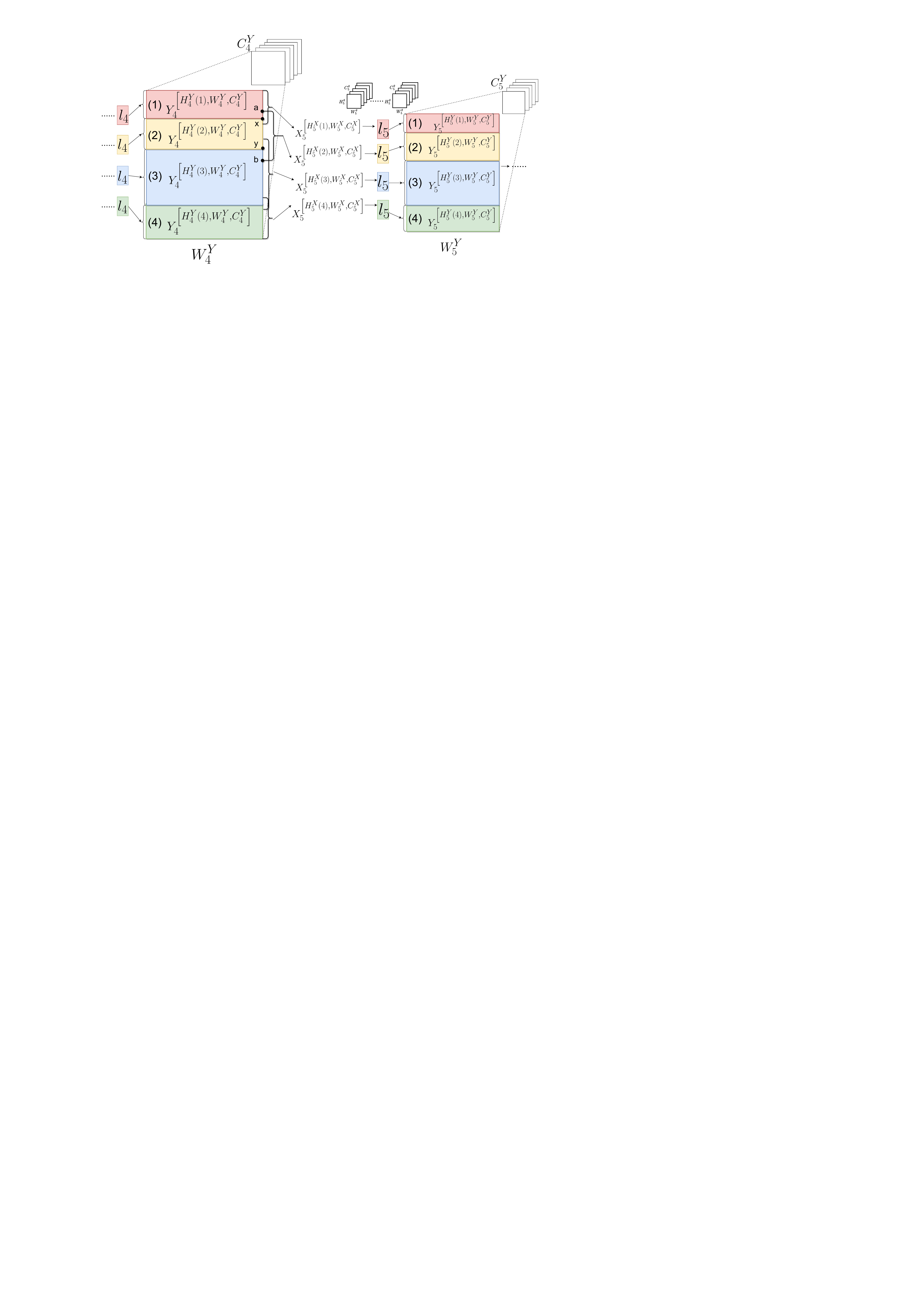}
	\caption{Data Partitioning Example}
	\label{fig:data_partitioning_energy}
\end{figure*}

An example of the data partitioning strategy illustrated in Figure~\subref*{fig:data_partition} is shown in Figure~\ref{fig:data_partitioning_energy}.
\begin{table*}[!t]
\begin{tabular}{p{0.99\textwidth}}
\begin{equation}
\label{equ:Hcomm}
H^{comm}_i \left( j \right) =  \left\{\begin{matrix}
\begin{aligned}
&x\left ( j \right )-a\left ( j \right )\;\;\;\;\;\;\;\;\;\;\;\;\;\;\;\;\;\;\;\;\;\;\; if (a\left ( j \right )< x\left ( j \right ),y\left ( j \right )\geqslant b\left ( j \right ))\\ 
&x\left ( j \right )-a\left ( j \right )+b\left ( j \right )-y\left ( j \right ) if (a\left ( j \right )< x\left ( j \right ),y\left ( j \right )<b\left ( j \right ))\\ 
&b\left ( j \right )-y\left ( j \right )\;\;\;\;\;\;\;\;\;\;\;\;\;\;\;\;\;\;\;\;\;\; if (a\left ( j \right )\geqslant x\left ( j \right ),y\left ( j \right )< b\left ( j \right ))\\ 
&0 \;\;\;\;\;\;\;\;\;\;\;\;\;\;\;\;\;\;\;\;\;\;\;\;\;\;\;\;\;\;\;\;\;\;\;\;\;\;\;if (a\left ( j \right )\geqslant x\left ( j \right ),y\left ( j \right )\geqslant b\left ( j \right ))\\ 
\end{aligned}
\end{matrix}\right.
\end{equation}\\
\begin{minipage}{.5\linewidth}
\begin{equation}
x\left ( j \right )=\left\{\begin{matrix}
\begin{aligned}
&1 \;\;\;\;\;\;\;\;\;\;\;\;\;\;\;\;\;\;if \left ( j=1 \right )\\
&\sum_{k=1}^{j-1} H_i^Y\left ( k \right )\;\; if \left ( j> 1 \right )
\end{aligned}
\end{matrix}\right.
\end{equation}\\
\end{minipage}%
\begin{minipage}{.5\linewidth}
\begin{equation}
y\left ( j \right ) = x\left ( j \right ) + H_i^Y \left ( j \right )
\end{equation}\\
\end{minipage}%
\\
\begin{minipage}{.5\linewidth}
\begin{equation}
\begin{aligned}
a\left ( j \right ) = a\left ( j-1 \right ) +H_{i+1}^X \left ( j-1 \right ) - H_{i+1}^\theta + s_{i+1} =\\ 1- \left ( j-1 \right )\cdot\left ( H_{i+1}^\theta - s_{i+1} \right ) + \sum_{k=1}^{j-1} H_{i+1}^{X}\left ( k \right )
\end{aligned}
\end{equation}\\
\end{minipage}%
\begin{minipage}{.5\linewidth}
\begin{equation}
\label{equ:bj}
b\left ( j \right ) = a\left ( j \right ) + H_{i+1}^{X}\left ( j \right )
\end{equation}\\
\end{minipage}%
\end{tabular}
\end{table*}
Layer $l_4$ and layer $l_5$ are two consecutive layers. $Y_4^{\left [ H^Y_4 ,W^Y_4 ,C^Y_4 \right ]}$ is the output tensor of layer $l_4$, which is partitioned along the height into four tensors $Y_4^{\left [ H^Y_4 \left(1\right) ,W^Y_4 ,C^Y_4 \right ]}$, $Y_4^{\left [ H^Y_4 \left(2\right) ,W^Y_4 ,C^Y_4 \right ]}$, $Y_4^{\left [ H^Y_4 \left(3\right) ,W^Y_4 ,C^Y_4 \right ]}$, $Y_4^{\left [ H^Y_4 \left(4\right) ,W^Y_4 ,C^Y_4 \right ]}$ and $H^Y_4 = \sum_{j=1}^{4} H^Y_4 \left(j\right)$. $X_5^{\left [ H^X_5 ,W^X_5 ,C^X_5 \right ]}=Y_4^{\left [ H^Y_4 ,W^Y_4 ,C^Y_4 \right ]}$ is the input tensor of layer $l_5$, which is partitioned along the height into four tensors $X_5^{\left [ H^X_5 \left(1\right) ,W^X_5 ,C^X_5 \right ]}$, $X_5^{\left [ H^X_5 \left(2\right) ,W^X_5 ,C^X_5 \right ]}$, $X_5^{\left [ H^X_5 \left(3\right) ,W^X_5 ,C^X_5 \right ]}$, $X_5^{\left [ H^X_5 \left(4\right) ,W^X_5 ,C^X_5 \right ]}$. $Y_5^{\left [ H^Y_5 ,W^Y_5 ,C^Y_5 \right ]}$ is the output tensor of layer $l_5$, which is partitioned along the height into four tensors $Y_5^{\left [ H^Y_5 \left(1\right) ,W^Y_5 ,C^Y_5 \right ]}$, $Y_5^{\left [ H^Y_5 \left(2\right) ,W^Y_5 ,C^Y_5 \right ]}$, $Y_5^{\left [ H^Y_5 \left(3\right) ,W^Y_5 ,C^Y_5 \right ]}$, $Y_5^{\left [ H^Y_5 \left(4\right) ,W^Y_5 ,C^Y_5 \right ]}$ and $H^Y_5 = \sum_{j=1}^{4} H^Y_5 \left(j\right)$. 

In order to generate an output tensor $Y_i^{\left [ H^Y_i \left(j\right) ,W^Y_i ,C^Y_i \right ]}$ from every partition $j$ with height of data $H^Y_i \left(j\right)$, the height of the input tensor $X_i^{\left [ H^X_i\left(j\right) ,W^X_i ,C^X_i \right ]}$ to every partition $j$ is $H^X_i \left(j\right) = \left[H^Y_i \left(j\right)-1\right]\cdot s_i+H^\Theta_i$. Additionally, we define $k_i\left(j\right) = H^Y_i \left(j\right)/H^Y_i$ as the fraction of partition $j$ in terms of internal computation and communication workload from the entire layer workload. For CNN layers within the same partition, they need to communicate data with their predecessor and successor layers internally, thus the fraction of this internal data communication is also $k_i\left(j\right)$. For the same CNN layer distributed onto different partitions, the size of the data tensor, which is produced by the predecessor layer may be smaller than what is needed for generating the data tensor for the successor layer, so a partition needs to communicate externally with its neighboring partitions for data exchange. For example, in Figure~\ref{fig:data_partitioning_energy}, the size of the produced tensor $Y_4^{\left [ H^Y_4 \left(2\right) ,W^Y_4 ,C^Y_4 \right ]}$ is smaller than the size of the needed tensor $X_5^{\left [ H^X_5 \left(2\right) ,W^X_5 ,C^X_5 \right ]}$ because $H_{5}^{X}\left ( 2 \right )>H_{4}^{Y}\left ( 2 \right )$, so partition 2 needs to communicate data externally with partition 1 and partition 3. 

\begin{table*}[!t]
\begin{tabular}{p{0.99\textwidth}}
\begin{equation}
   E_j =E_{j}^{comp}+E_{j}^{in,comm}+E_{j}^{ex,comm}
   \label{equ:energy_sum}
\end{equation}\\
\begin{equation}
	E_{j}^{comp} =\sum_{\forall l_i \in L} k_i\left(j\right)\cdot \int_{0}^{t_i} P_{i} \left ( t \right ) dt = \sum_{\forall l_i \in L} \left(H^Y_i \left(j\right)/H^Y_i \right)\cdot  \int_{0}^{t_i} P_{i} \left ( t \right ) dt
\end{equation}\\
\begin{equation}
	E_{j}^{in,comm} = \sum_{\forall l_i \in L} k_i\left(j\right)\cdot E_{in,comm}^{i}\left(H^Y_i \times W^Y_i  \times C^Y_i \right) 
	=\sum_{\forall l_i \in L} \left(H^Y_i \left(j\right)/H^Y_i \right)\cdot E_{in,comm}^{i}\left(H^Y_i \times W^Y_i  \times C^Y_i \right)
\end{equation}\\
\begin{equation}
\label{equ:Ejexcomm}
	E_{j}^{ex,comm} = \sum_{\forall l_i \in L \setminus l_q} \left(H^{comm}_i \left(j\right) /H^Y_i \right)\cdot E_{ex,comm}^{i}\left(H^Y_i \times W^Y_i  \times C^Y_i \right)
\end{equation}\\
\end{tabular}
\end{table*}

The height of the externally communicated data tensor $H^{comm}_i \left( j \right)$ depends on the relative location of the produced piece of data and needed piece of data. Considering Figure~\ref{fig:data_partitioning_energy}, we denote the starting point and ending point in height of the produced piece of data as $x\left ( j \right )$ and $y\left ( j \right )$, respectively. The starting point and ending point in height of the needed piece of data is denoted as $a\left ( j \right )$ and $b\left ( j \right )$, respectively, as shown in Figure~\ref{fig:data_partitioning_energy}. Using the aforementioned points $x\left ( j \right )$, $y\left ( j \right )$, $a\left ( j \right )$, and $b\left ( j \right )$ for each partition $j$, we can define four cases and we can calculate $H^{comm}_i\left( j \right)$ as shown in Equation~\ref{equ:Hcomm} to Equation~\ref{equ:bj}.

Based on the above discussion, we devise our analytical energy consumption model for the data partitioning strategy as shown in Equation~\ref{equ:energy_sum} to Equation~\ref{equ:Ejexcomm}. Energy $E_j$ is the energy consumed by partition $j$ when it is executed on an edge device in a distributed system at the edge. $E_j$ consists of three main parts, as given in Equation \ref{equ:energy_sum}. 

Energy $E_{j}^{comp}$ is the energy consumed for the CNN inference computation in partition $j$. $E_{j}^{in,comm}$ is the energy consumed for the internal data communication in partition $j$. $E_{j}^{ex,comm}$ is the energy consumed for the external data communication between partition $j$ and neighboring partitions. Time $t_i$ is the execution time of layer $l_i$ on an edge device to generate the whole output data tensor $Y_i^{\left [ H^Y_i, W^Y_i, C^Y_i \right ]}$. $P_{i}\left(t\right)$ is the power consumption when layer $l_i$ is executed on an edge device to generate the whole output data tensor. Both $t_i$ and $P_{i}\left(t\right)$ can be acquired by CNN layer profiling on the edge device. $E_{in,comm}^{i}\left(H^Y_i \times W^Y_i  \times C^Y_i \right)$ is the energy consumption for layer $l_i$ to communicate internally within the same edge device the whole output data tensor $Y_i^{\left [ H^Y_i, W^Y_i, C^Y_i \right ]}$. $ E_{ex,comm}^{i}\left(H^Y_i \times W^Y_i  \times C^Y_i \right)$ is the energy consumption for layer $l_i$ to communicate externally data tensor $Y_i^{\left [ H^Y_i, W^Y_i, C^Y_i \right ]}$ across edge devices in a distributed system, that can be acquired by profiling of the external data communication protocol implemented and executed on an edge device. $l_q$ is the last layer in the entire CNN model.

\subsection{Energy Model for Horizontal Partitioning Strategy}
\label{sec:model_horizontal}
\begin{figure*}[!t]
	\includegraphics[width=.8\linewidth]{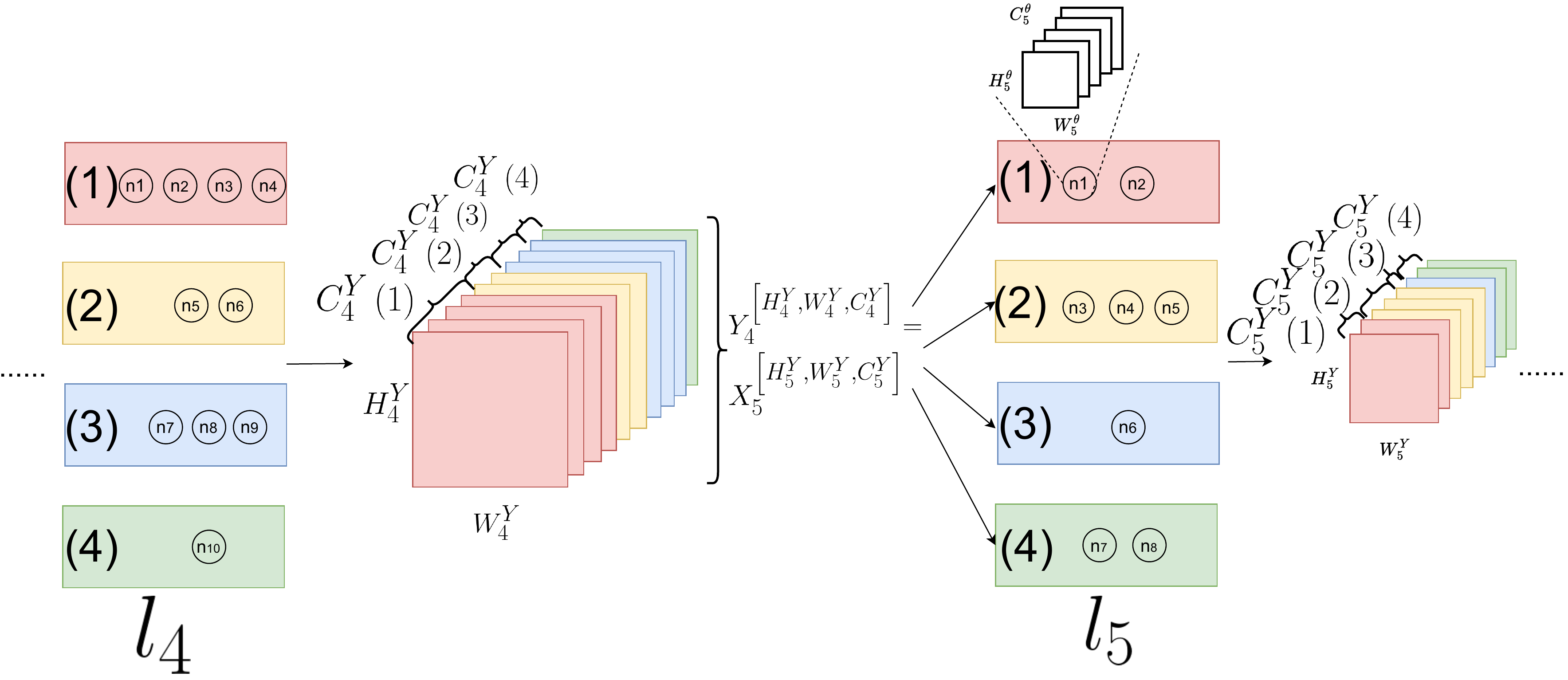}
	\caption{Horizontal Partitioning Example}
	\label{fig:horizontal_partitioning_energy}
\end{figure*}

For the horizontal partitioning strategy, the input data given to a CNN layer $l_i$ is not partitioned whereas the weights $\Theta _{i}^{\left [ N_{i}^{\Theta}, H_{i}^{\Theta}, W_{i}^{\Theta}, C_{i}^{\Theta} \right ]}$ of every CNN layer are partitioned by grouping neurons into $M$ partitions in such a way that the weights $\Theta _{i}^{\left [ N_{i}^{\Theta}\left ( j \right ), H_{i}^{\Theta}, W_{i}^{\Theta}, C_{i}^{\Theta} \right ]}$ for every partition $j$ correspond to the group of neurons in partition $j$. This means that $N_i^{\Theta} = \sum_{j=1}^{M} N_i^{\Theta} \left(j\right)$. An example of the horizontal partitioning strategy illustrated in Figure~\subref*{fig:horizental_partition} is shown in Figure~\ref{fig:horizontal_partitioning_energy}.
\begin{table*}[!t]
\begin{tabular}{p{0.99\textwidth}}
\begin{equation}
E_j =E_{j}^{comp}+E_{j}^{in,comm}+E_{j}^{ex,comm}
\label{equ:energy_sum1}
\end{equation}\\
\begin{equation}
E_{j}^{comp} =\sum_{\forall l_i \in L} k_i\left(j\right)\cdot \int_{0}^{t_i} P_{i} \left ( t \right ) dt = \sum_{\forall l_i \in L} \left(C_i^Y \left(j\right)/C_i^Y \right)\cdot \int_{0}^{t_i} P_{i} \left ( t \right ) dt 
\end{equation}\\
\begin{equation}	
E_{j}^{in,comm} = \sum_{\forall l_i \in L} k_i\left(j \right)\cdot E_{in,comm}^{i}\left(H^Y_i \times W^Y_i  \times C^Y_i \right) =\sum_{\forall l_i \in L} \left(C_i^Y \left(j\right)/C_i^Y \right)\cdot E_{in,comm}^{i}\left(H^Y_i \times W^Y_i  \times C^Y_i \right)
\end{equation}\\
\begin{equation}
\label{equ:Ejexcomm1}
E_{j}^{ex,comm} = \sum_{\forall l_i \in L\setminus l_q} \left(1-C_i^Y \left(j\right)/C_i^Y \right)\cdot E_{ex,comm}^{i} \left(H^Y_i \times W^Y_i  \times C^Y_i \right)
\end{equation}\\
\end{tabular}
\end{table*}
Layers $l_4$ and $l_5$ are two consecutive layers. The weights $\Theta_4^{\left [ 10, H_{4}^{\Theta}, W_{4}^{\Theta}, C_{4}^{\Theta} \right ]}$ of layer $l_4$ are divided into four partitions by grouping the ten neurons of $l_4$ into the following four groups: $\left \{ n_1, n_2, n_3, n_4 \right \}$, $\left \{ n_5, n_6\right \}$, $\left \{ n_7, n_8, n_9 \right \}$, and $\left \{ n_{10} \right \}$. This grouping results in the following four partitions of weights: $\Theta _{4}^{\left [ N_{4}^{\Theta}\left ( 1 \right ), H_{4}^{\Theta}, W_{4}^{\Theta}, C_{4}^{\Theta} \right ]}$, $\Theta _{4}^{\left [ N_{4}^{\Theta}\left ( 2 \right ), H_{4}^{\Theta}, W_{4}^{\Theta}, C_{4}^{\Theta} \right ]}$, $\Theta _{4}^{\left [ N_{4}^{\Theta}\left ( 3 \right ), H_{4}^{\Theta}, W_{4}^{\Theta}, C_{4}^{\Theta} \right ]}$, and $\Theta _{4}^{\left [ N_{4}^{\Theta}\left ( 4 \right ), H_{4}^{\Theta}, W_{4}^{\Theta}, C_{4}^{\Theta} \right ]}$, where $N_{4}^{\Theta }\left ( 1 \right )=4$, $N_{4}^{\Theta }\left ( 2 \right )=2$, $N_{4}^{\Theta }\left ( 3 \right )=3$, and $N_{4}^{\Theta }\left ( 4 \right )=1$. Analogously, the weights $\Theta _{5}^{\left [ 8, H_{5}^{\Theta}, W_{5}^{\Theta}, C_{5}^{\Theta} \right ]}$ of layer $l_5$ are divided into four partitions of weights: $\Theta _{5}^{\left [ 2, H_{5}^{\Theta}, W_{5}^{\Theta}, C_{5}^{\Theta} \right ]}$, $\Theta _{5}^{\left [ 3, H_{5}^{\Theta}, W_{5}^{\Theta}, C_{5}^{\Theta} \right ]}$, $\Theta _{5}^{\left [ 1, H_{5}^{\Theta}, W_{5}^{\Theta}, C_{5}^{\Theta} \right ]}$, and $\Theta _{5}^{\left [ 2, H_{5}^{\Theta}, W_{5}^{\Theta}, C_{5}^{\Theta} \right ]}$, where partition 1 includes neurons $\left \{ n_1, n_2 \right \}$, partition 2 includes neurons $\left \{ n_3, n_4, n_5 \right \}$, partition 3 includes neurons $\left \{ n_6 \right \}$, and partition 4 includes neurons $\left \{ n_7, n_8 \right \}$. 

Every partition $j$ of layer $l_i$ takes the whole input data tensor $X_i^{\left [ H^X_i ,W^X_i ,C^X_i \right ]}$ and generates only part of the output data tensor, denoted as $Y_i^{\left [ H^Y_i ,W^Y_i ,C^Y_i \left( j \right)  \right ]}$. For example, partition 2 of layer $l_5$ in Figure \ref{fig:horizontal_partitioning_energy} takes $X_5^{\left [ H^X_5 ,W^X_5 ,C^X_5 \right ]}$ and generates part of the output data tensor $Y_5^{\left [ H^Y_5 ,W^Y_5 ,C^Y_5 \left( 2 \right) \right ]}$ where $C_{5}^{Y}\left ( 2 \right )=3$. We define $k_i\left(j\right) = C_i^Y \left(j\right)/C_i^Y$ as the fraction of partition $j$ in terms of internal computation and communication workload from the entire layer workload. For CNN layers within the same partition $j$, they need to communicate data internally with their predecessor and successor layers, thus the fraction of this internal communication is also $k_i\left(j\right)$. Every partition $j$ of layer $l_{i+1}$ needs to communicate externally with all other partitions $p\neq j$ of layer $l_i$ because every partition $j$ of layer $l_{i+1}$ needs the whole input data tensor $X_{i+1}^{\left [ H_{i+1}^{X}, W_{i+1}^{X}, C_{i+1}^{X} \right ]}=Y_{i}^{\left [ H_{i}^{Y}, W_{i}^{Y}, C_{i}^{Y} \right ]}$. For example, in Figure \ref{fig:horizontal_partitioning_energy}, partition 3 of layer $l_5$ needs to communicate externally with partitions 1, 2, and 4 of layer $l_4$ in order to obtain data tensors $Y_4^{\left [ H^Y_4 ,W^Y_4 ,C^Y_4 \left( 1 \right) \right ]}$, $Y_4^{\left [ H^Y_4 ,W^Y_4 ,C^Y_4 \left( 2 \right) \right ]}$, and $Y_4^{\left [ H^Y_4 ,W^Y_4 ,C^Y_4 \left( 4 \right) \right ]}$. 

Based on the above discussion, we devise our analytical energy consumption model for the horizontal partitioning strategy as shown in Equation~\ref{equ:energy_sum1} to Equation~\ref{equ:Ejexcomm1}. Energy $E_j$ is the energy consumed by partition $j$ when it is executed on an edge device in a distributed system at the edge. $E_j$ consists of three main parts, as given in Equation~\ref{equ:energy_sum1}.

Energy $E_{j}^{comp}$ is the energy consumed for the CNN inference computation in partition $j$. $E_{j}^{in,comm}$ is the energy consumed for the internal communication in partition $j$. $E_{j}^{ex,comm}$ is the energy consumed for the external data communication between partition $j$ and all the other partitions. Time $t_i$ is the execution time of layer $l_i$ on an edge device to generate the whole output data tensor $Y_i^{\left [ H^Y_i, W^Y_i, C^Y_i \right ]}$. $P_{i}\left(t\right)$ is the power consumption when layer $l_i$ is executed on an edge device to generate the whole output data tensor. Both $t_i$ and $P_{i}\left(t\right)$ can be acquired by CNN layer profiling on the edge device. $E_{in,comm}^{i}\left(H^Y_i \times W^Y_i  \times C^Y_i \right)$ is the energy consumption for layer $l_i$ to communicate internally within the same edge device, the whole output data tensor $Y_i^{\left [ H^Y_i, W^Y_i, C^Y_i \right ]}$. $ E_{ex,comm}^{i}\left(H^Y_i \times W^Y_i  \times C^Y_i \right)$ is the energy consumption for layer $l_i$ to communicate externally data tensor $Y_i^{\left [ H^Y_i, W^Y_i, C^Y_i \right ]}$ across edge devices in a distributed system, that can be acquired by profiling. $l_q$ is the last layer in the entire CNN model.

\subsection{Energy Model for Sequential Partitioning Strategy}
\label{sec:model_sequential}
For the sequential partitioning strategy, the layers of a CNN model are grouped in $M$ partitions and each partition $p_j$ includes consecutive CNN layers. The input data, given to a CNN layer, and the layer's weights are not partitioned. An example of the sequential partitioning strategy is shown in Figure~\subref*{fig:sequantial_partition}. The entire CNN model is divided into $M=4$ partitions by grouping the ten layers of the CNN model into the following four groups: $p_1=\left \{ l_1, l_2 \right \}$, $p_2=\left \{ l_3, l_4, l_5 \right \}$, $p_3=\left \{ l_6, l_7 \right \}$, and $p_4=\left \{ l_8, l_9, l_{10} \right \}$. 

\begin{table*}[!t]
\begin{tabular}{p{0.99\textwidth}}
\begin{equation}
   E_j = E_j^{comp}+E_j^{in,comm}+E_j^{ex,comm}
   \label{equ:energy_sum2}
\end{equation}\\ 
\begin{equation}
  E_j^{comp} =\sum_{\forall l_i \in p_j}\int_{0}^{t_i} P_{i} \left ( t \right ) dt 
\end{equation}\\
\begin{equation}
  E_j^{in,comm} = \sum_{\forall l_i \in p_j} E_{in,comm}^{i}\left(H^Y_i \times W^Y_i  \times C^Y_i \right) 
\end{equation}\\
\begin{equation}
\label{equ:Ejexcomm2}
E_{j}^{ex,comm} =  \left\{\begin{matrix}
&E_{ex,send}^{b}\left (  H^Y_b  \times W^Y_b \times C^Y_b  \right )\;\;\;\;\;\;\;\;\;\;\;\;\;\;\;\;\;\;\;\;\;\;~~~~~~~~~~~~~~~~~~~~~~~~~~~~~~~~~~~~~~~~ if \ l_z\in p_j\\ 
&E_{ex,receive}^{a}\left (  H^X_a  \times W^X_a \times C^X_a  \right )\;\;\;\;\;\;\;\;\;\;\;\;\;\;\;\;\;\;~~~~~~~~~~~~~~~~~~~~~~~~~~~~~~~~~~~~~~~~ if \ l_q\in p_j\\ 
&E_{ex,receive}^{a}\left (  H^X_a  \times W^X_a \times C^X_a  \right ) + E_{ex,send}^{b}\left (  H^Y_b  \times W^Y_b \times C^Y_b  \right ) ~~~~~~~otherwise
\end{matrix}\right.
\end{equation}\\
\end{tabular}
\end{table*}

CNN layers within the same partition $p_j$ need to communicate internally with their predecessor and successor layers (if they exist). Each partition $p_j$ in a distributed system only needs to communicate externally with its predecessor and successor partitions (the first partition only needs to communicate with its successor partition and the last partition only needs to communicate with its predecessor partition). For example, layer $l_4$ of partition $p_2$ in Figure~\subref*{fig:sequantial_partition} needs to communicate internally with layers $l_3$ and $l_5$. Partition $p_2$ needs to communicate externally with partition $p_1$ and partition $p_3$.

Based on the above discussion, we devise our analytical energy consumption model for the sequential partitioning strategy as shown in Equation~\ref{equ:energy_sum2} to Equation~\ref{equ:Ejexcomm2}. Energy $E_j$ is the energy consumed by partition $p_j$ when it is executed on an edge device in a distributed system at the edge. $E_j$ consists of three main parts, as given in Equation \ref{equ:energy_sum2}. Energy $E_{j}^{comp}$ is the energy consumed for the CNN inference computation in partition $p_j$. $E_{j}^{in,comm}$ is the energy consumed for the internal communication in partition $p_j$. $E_{j}^{ex,comm}$ is the energy consumed for the external communication between partition $p_j$ and its predecessor and successor partitions. Time $t_i$ is the execution time of layer $l_i$ on an edge device to generate the whole output data tensor $Y_i^{\left [ H^Y_{i}, W^Y_i, C^Y_i \right ]}$. $P_{i}\left(t\right)$ is the power consumption when layer $l_i$ is executed on an edge device to generate the whole output data tensor. Both $t_i$ and $P_{i}\left(t\right)$ can be acquired by CNN layer profiling on the edge device. $E_{in,comm}^{i}\left(H^Y_i \times W^Y_i  \times C^Y_i \right) $ is the energy consumption for layer $l_i$ to communicate internally within the same edge device, the whole output data tensor $Y_i^{\left [ H^Y_i, W^Y_i, C^Y_i \right ]}$. $l_z$ is the first layer in the entire CNN model and $l_q$ is the last layer in the entire CNN model. $E_{ex,receive}^{a}\left (  H^X_a  \times W^X_a \times C^X_a  \right )$ is the energy consumption for the first layer $l_a$ in partition $p_j$ to receive input data tensor $X_a^{\left [ H^X_a, W^X_a, C^X_a \right ]}$ from another edge device in a distributed system, that can be acquired by profiling. $E_{ex,send}^{b}\left (  H^Y_b  \times W^Y_b \times C^Y_b  \right )$ is the energy consumption for the last layer $l_b$ in partition $p_j$ to send externally output data tensor $Y_b^{\left [ H^Y_b, W^Y_b, C^Y_b \right ]}$ to another edge device in a distributed system, that can be acquired by profiling.

\subsection{Energy Model for Vertical Partitioning Strategy}
\label{sec:model_vertical}
For the vertical partitioning strategy, the layers of a CNN model are grouped in $M$ partitions and each partition $p_j$ may include non-consecutive CNN layers. The input data, given to a CNN layer, and the layer's weights are not partitioned. An example of the vertical partitioning strategy is shown in Figure~\subref*{fig:vertical_partition}. The entire CNN model is divided into four partitions by grouping the ten layers of the CNN model into the following four groups: $p_1=\left \{ l_1, l_2, l_8 \right \}$, $p_2=\left \{ l_3, l_9, l_{10} \right \}$, $p_3=\left \{ l_4, l_5, l_7 \right \}$, and $p_4=\left \{ l_6 \right \}$. 

The CNN layers within the same partition $p_j$ can be divided into $m_j$ sub-partitions where each sub-partition $p_{js}$ includes consecutive CNN layers. For example, in Figure \subref*{fig:vertical_partition}, partition $p_1$ can be divided into two sub-partitions $p_{11}$ and $p_{12}$ where sub-partition $p_{11}$ includes layers $l_1$ and $l_2$ and sub-partition $p_{12}$ includes layer $l_8$. CNN layers within the same sub-partition $p_{js}$ need to communicate internally with their predecessor and successor layers (if they exist). For example, layer $l_1$ of sub-partition $p_{11}$ in Figure~\subref*{fig:vertical_partition} needs to communicate internally with layer $l_{2}$. Each sub-partition $p_{js}$ in a distributed system only needs to communicate externally with its predecessor and successor sub-partitions. For example, sub-partition $p_{12}$ in Figure~\subref*{fig:vertical_partition} needs to communicate externally with sub-partition $p_{32}$ and sub-partition $p_{22}$. 

Based on the above discussion, we devise our analytical energy consumption model for the vertical partitioning strategy as shown in Equation~\ref{equ:energy_sum3} to Equation~\ref{equ:Ejs}. 
\begin{table*}[!t]
\begin{tabular}{p{0.99\textwidth}}
\begin{equation}
E_j = E_j^{comp}+E_j^{in,comm}+E_j^{ex,comm}
\label{equ:energy_sum3}
\end{equation}\\ 
\begin{equation}
E_j^{comp} =\sum_{\forall l_i \in p_j}\int_{0}^{t_i} P_{i} \left ( t \right ) dt 
\end{equation}\\
\begin{equation}
E_j^{in,comm} = \sum_{s=1}^{m_j}\sum_{\forall l_i \in p_{js}} E_{in,comm}^{i}\left(H^Y_i \times W^Y_i  \times C^Y_i \right) 
\end{equation}\\
\begin{equation}
E_{j}^{ex,comm} = \sum_{s=1}^{m_j} E_{js}^{ex,comm}
\end{equation}\\
\begin{equation}
\label{equ:Ejs}
E_{js}^{ex,comm} = \left\{\begin{matrix}
&E_{ex,send}^{b}\left (  H^Y_b  \times W^Y_b \times C^Y_b  \right )\;\;\;\;\;\;\;\;\;\;\;\;\;\;\;\;\;\;~~~~~~~~~~~~~~~~~~~~~~~~~~~~~~~~~~~~~~~~~~ if \ l_z\in p_{js}\\ &E_{ex,receive}^{a}\left (  H^X_a  \times W^X_a \times C^X_a  \right )\;\;\;\;\;\;\;\;\;\;\;\;\;\;~~~~~~~~~~~~~~~~~~~~~~~~~~~~~~~~~~~~~~~~~~ if \ l_q\in p_{js}\\ 
&E_{ex,receive}^{a}\left (  H^X_a  \times W^X_a \times C^X_a  \right ) + E_{ex,send}^{b}\left (  H^Y_b  \times W^Y_b \times C^Y_b  \right )~~~~~otherwise
\end{matrix}\right.
\end{equation}\\
\end{tabular}
\end{table*}
Energy $E_j$ is the energy consumed by partition $p_j$ executed on an edge device in a distributed system at the edge. $E_j$ consists of three main parts, as given in Equation \ref{equ:energy_sum3}. $E_{j}^{comp}$ is the energy consumed for the CNN inference computation in partition $p_j$. $E_{j}^{in,comm}$ is the energy consumed for the internal communication in partition $p_j$. $E_{j}^{ex,comm}$ is the energy consumed for the external communication between partition $p_j$ and other partitions. Time $t_i$ is the execution time of layer $l_i$ on an edge device to generate the whole output data tensor $Y_i^{\left [ H^Y_{i}, W^Y_i, C^Y_i \right ]}$. $P_{i}\left(t\right)$ is the power consumption when layer $l_i$ is executed on an edge device to generate the whole output data tensor. Both $t_i$ and $P_{i}\left(t\right)$ can be acquired by CNN layer profiling on the edge device. $E_{in,comm}^{i}\left(H^Y_i \times W^Y_i  \times C^Y_i \right) $ is the energy consumption for layer $l_i$ to communicate internally within the same edge devices, the whole output data tensor $Y_i^{\left [ H^Y_i, W^Y_i, C^Y_i \right ]}$. $l_z$ is the first layer in the entire CNN model and $l_q$ is the last layer in the entire CNN model. $E_{ex,receive}^{a}\left (  H^X_{a}  \times W^X_{a} \times C^X_{a}  \right )$ is the energy consumption for the first layer $l_{a}$ in sub-partition $p_{js}$ to receive input data tensor $X_{a}^{\left [ H^X_{a}, W^X_{a}, C^X_{a} \right ]}$ from another edge device in a distributed system, that can be acquired by profiling. $E_{ex,send}^{b}\left ( H^Y_{b}  \times W^Y_{b} \times C^Y_{b}  \right )$ is the energy consumption for the last layer $l_{b}$ in sub-partition $p_{js}$ to send externally output data tensor $Y_{b}^{\left [ H^Y_{b}, W^Y_{b}, C^Y_{b} \right ]}$ to another edge device in a distributed system, that can be acquired by profiling.
\section{Validation of Energy Consumption Models}
\label{sec:validation}
In this section, we validate our energy consumption analytical models, introduced in Section~\ref{sec:model}, for the four partitioning strategies that enable distributed CNN model inference at the edge. The goal of this validation is to demonstrate that our energy consumption analytical models are sufficiently accurate to compare the energy consumption of the different partitioning strategies such that we can confidently use the models in the large-scale energy consumption investigation, presented in Section~\ref{sec:investigation}.  First, we explain the setup for our model validation experiments in Section~\ref{sec:Experimental_setup}. Then, in Section~\ref{sec:accuracy}, we present the experimental results and confirm the accuracy of our analytical models.

\subsection{Experimental setup}
\label{sec:Experimental_setup}
We experiment with two real-world CNNs, namely VGG 16 and Emotion\_fer mentioned in Section~\ref{sec:CNNs} - Table~\ref{tab:CNNs}, that take images as input for CNN inference. These CNNs are utilized in many different applications and have diverse number of layers, input data size, and weights size. Such a diversity leads to a diverse scale of energy consumption when these CNNs are mapped and executed on one and the same distributed system at the edge by using the same partitioning strategy. VGG 16 is used for image classification and has 23 layers. Emotion\_fer is used for body, face, and gesture analysis and has 19 layers. So, these two CNN models are sufficiently representative and good examples to apply our energy consumption analytical models for the four partitioning strategies and to demonstrate the energy models accuracy. The two CNN models, mentioned above, are partitioned, mapped, and executed on a homogeneous distributed system at the edge which consists of four NVIDIA Jetson TX2 embedded platforms that communicate through MPI, as mentioned in Section~\ref{sec:Distributed_hardware_platform}. We utilize the four partitioning strategies on the VGG 16 model to obtain sixteen partitioned models. That is, every partitioning strategy is utilized to obtain four partitioned VGG 16 models: one model with one partition, one model with two partitions, one model with three partitions, and one model with four partitions. The number of partitions in a model determines the number of NVIDIA Jetson TX2 embedded platforms that are used to execute the partitions on the distributed system, i.e., every partition is executed on a separate Jetson TX2 platform.    

\newfloatcommand{capbtabbox}{table}[][\FBwidth]
	\begin{figure*}
		\hspace{-3cm}
		\begin{floatrow}
			\ffigbox{%
				\rule{0cm}{6cm}%
			}{%
			   \includegraphics[width=0.62\linewidth]{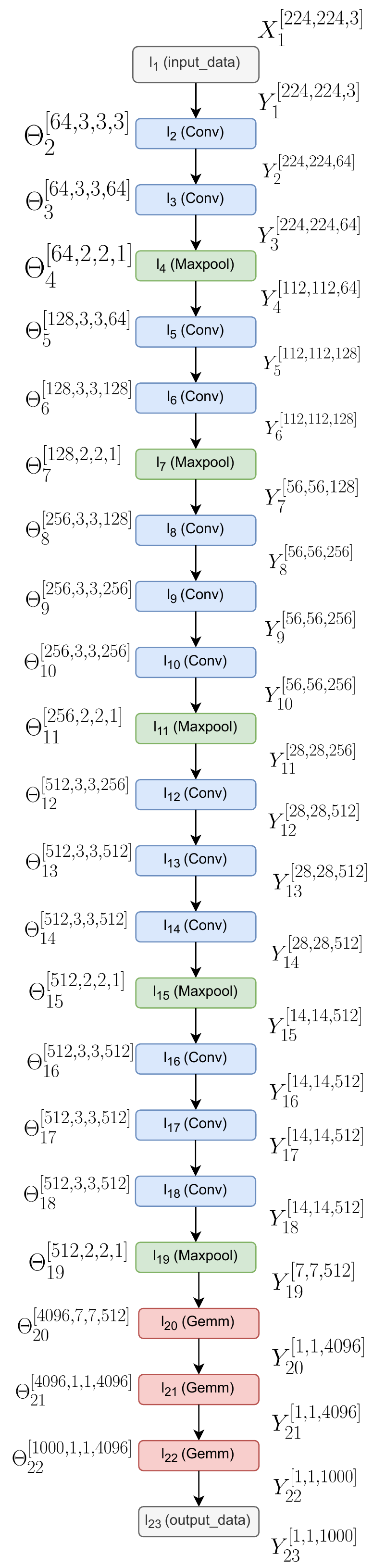}
			   \caption{VGG 16 network structure}%
			   \label{fig:VGG16_network_structure}
			}
			\capbtabbox{%
				\setlength{\tabcolsep}{10pt}
                \resizebox{0.55\textwidth}{!}{%
                	\hspace{-4cm}
					\begin{tabular}{|c|ccc|ccc|ccccccc|ccccccc|}
                      \hline
                      \multicolumn{1}{|c|}{\multirow{3}{*}{layer}} & \multicolumn{4}{c|}{Data Partitioning}                                                                                                                                                                                                                                                                                                                                                                    & \multicolumn{4}{c|}{Horizontal Partitioning}                                                                                                                                                                                                                                                                                                                                                                                                                              \\ \cmidrule(l){2-9} 
                      \multicolumn{1}{|c|}{}                       & \multicolumn{4}{c|}{number of partitions}                                                                                                                                                                                                                                                                                                                                                                 & \multicolumn{4}{c|}{number of partitions}                                                                                                                                                                                                                                                                                                                                                                                                                                 \\ \cmidrule(l){2-9} 
                      \multicolumn{1}{|c|}{}                       & \multicolumn{1}{c|}{1}                  & \multicolumn{1}{c|}{2}                                                                           & \multicolumn{1}{c|}{3}                                                                                              & \multicolumn{1}{c|}{4}                                                                                                                 & \multicolumn{1}{c|}{1}                       & \multicolumn{1}{c|}{2}                                                                                         & \multicolumn{1}{c|}{3}                                                                                                                   & \multicolumn{1}{c|}{4}                                                                                                                                         \\ \midrule
                      \multicolumn{1}{|c|}{$l_1$}                  & \multicolumn{1}{c|}{$H_1^Y(1)=224$}     & \multicolumn{1}{c|}{\begin{tabular}[c]{@{}c@{}}$H_1^Y(1)=112$\\ $H_1^Y(2)=112$\end{tabular}}     & \multicolumn{1}{c|}{\begin{tabular}[c]{@{}c@{}}$H_1^Y(1)=75$\\ $H_1^Y(2)=75$\\ $H_1^Y(3)=74$\end{tabular}}          & \multicolumn{1}{c|}{\begin{tabular}[c]{@{}c@{}}$H_1^Y(1)=56$\\ $H_1^Y(2)=56$\\ $H_1^Y(3)=56$\\ $H_1^Y(4)=56$\end{tabular}}             & \multicolumn{1}{c|}{N.A.}                    & \multicolumn{1}{c|}{N.A.}                                                                                      & \multicolumn{1}{c|}{N.A.}                                                                                                                & \multicolumn{1}{c|}{N.A.}                                                                                                                                      \\ \midrule
                      \multicolumn{1}{|c|}{$l_2$}                  & \multicolumn{1}{c|}{$H_2^Y(1)=224$}     & \multicolumn{1}{c|}{\begin{tabular}[c]{@{}c@{}}$H_2^Y(1)=112$\\ $H_2^Y(2)=112$\end{tabular}}     & \multicolumn{1}{c|}{\begin{tabular}[c]{@{}c@{}}$H_2^Y(1)=75$\\ $H_2^Y(2)=74$\\ $H_2^Y(3)=75$\end{tabular}}          & \multicolumn{1}{c|}{\begin{tabular}[c]{@{}c@{}}$H_2^Y(1)=56$\\ $H_2^Y(2)=56$\\ $H_2^Y(3)=56$\\ $H_2^Y(4)=56$\end{tabular}}             & \multicolumn{1}{c|}{$N_2^\Theta(1)=64$}      & \multicolumn{1}{c|}{\begin{tabular}[c]{@{}c@{}}$N_2^\Theta(1)=32$\\ $N_2^\Theta(2)=32$\end{tabular}}           & \multicolumn{1}{c|}{\begin{tabular}[c]{@{}c@{}}$N_2^\Theta(1)=22$\\ $N_2^\Theta(2)=21$\\ $N_2^\Theta(3)=21$\end{tabular}}                & \multicolumn{1}{c|}{\begin{tabular}[c]{@{}c@{}}$N_2^\Theta(1)=16$\\ $N_2^\Theta(2)=16$\\ $N_2^\Theta(3)=16$\\ $N_2^\Theta(4)=16$\end{tabular}}                 \\ \midrule
                      \multicolumn{1}{|c|}{$l_3$}                  & \multicolumn{1}{c|}{$H_3^Y(1)=224$}     & \multicolumn{1}{c|}{\begin{tabular}[c]{@{}c@{}}$H_3^Y(1)=112$\\ $H_3^Y(2)=112$\end{tabular}}     & \multicolumn{1}{c|}{\begin{tabular}[c]{@{}c@{}}$H_3^Y(1)=74$\\ $H_3^Y(2)=75$\\ $H_3^Y(3)=75$\end{tabular}}          & \multicolumn{1}{c|}{\begin{tabular}[c]{@{}c@{}}$H_3^Y(1)=56$\\ $H_3^Y(2)=56$\\ $H_3^Y(3)=56$\\ $H_3^Y(4)=56$\end{tabular}}             & \multicolumn{1}{c|}{$N_3^\Theta(1)=64$}      & \multicolumn{1}{c|}{\begin{tabular}[c]{@{}c@{}}$N_3^\Theta(1)=32$\\ $N_3^\Theta(2)=32$\end{tabular}}           & \multicolumn{1}{c|}{\begin{tabular}[c]{@{}c@{}}$N_3^\Theta(1)=21$\\ $N_3^\Theta(2)=22$\\ $N_3^\Theta(3)=21$\end{tabular}}                & \multicolumn{1}{c|}{\begin{tabular}[c]{@{}c@{}}$N_3^\Theta(1)=16$\\ $N_3^\Theta(2)=16$\\ $N_3^\Theta(3)=16$\\ $N_3^\Theta(4)=16$\end{tabular}}                 \\ \midrule
                      \multicolumn{1}{|c|}{$l_4$}                  & \multicolumn{1}{c|}{$H_4^Y(1)=112$}     & \multicolumn{1}{c|}{\begin{tabular}[c]{@{}c@{}}$H_4^Y(1)=56$\\ $H_4^Y(2)=56$\end{tabular}}       & \multicolumn{1}{c|}{\begin{tabular}[c]{@{}c@{}}$H_4^Y(1)=38$\\ $H_4^Y(2)=37$\\ $H_4^Y(3)=37$\end{tabular}}          & \multicolumn{1}{c|}{\begin{tabular}[c]{@{}c@{}}$H_4^Y(1)=28$\\ $H_4^Y(2)=28$\\ $H_4^Y(3)=28$\\ $H_4^Y(4)=28$\end{tabular}}             & \multicolumn{1}{c|}{$N_4^\Theta(1)=64$}      & \multicolumn{1}{c|}{\begin{tabular}[c]{@{}c@{}}$N_4^\Theta(1)=32$\\ $N_4^\Theta(2)=32$\end{tabular}}           & \multicolumn{1}{c|}{\begin{tabular}[c]{@{}c@{}}$N_4^\Theta(1)=21$\\ $N_4^\Theta(2)=21$\\ $N_4^\Theta(3)=22$\end{tabular}}                & \multicolumn{1}{c|}{\begin{tabular}[c]{@{}c@{}}$N_4^\Theta(1)=16$\\ $N_4^\Theta(2)=16$\\ $N_4^\Theta(3)=16$\\ $N_4^\Theta(4)=16$\end{tabular}}                 \\ \midrule
                      \multicolumn{1}{|c|}{$l_5$}                  & \multicolumn{1}{c|}{$H_5^Y(1)=112$}     & \multicolumn{1}{c|}{\begin{tabular}[c]{@{}c@{}}$H_5^Y(1)=56$\\ $H_5^Y(2)=56$\end{tabular}}       & \multicolumn{1}{c|}{\begin{tabular}[c]{@{}c@{}}$H_5^Y(1)=37$\\ $H_5^Y(2)=38$\\ $H_5^Y(3)=37$\end{tabular}}          & \multicolumn{1}{c|}{\begin{tabular}[c]{@{}c@{}}$H_5^Y(1)=28$\\ $H_5^Y(2)=28$\\ $H_5^Y(3)=28$\\ $H_5^Y(4)=28$\end{tabular}}             & \multicolumn{1}{c|}{$N_5^\Theta(1)=128$}     & \multicolumn{1}{c|}{\begin{tabular}[c]{@{}c@{}}$N_5^\Theta(1)=64$\\ $N_5^\Theta(2)=64$\end{tabular}}           & \multicolumn{1}{c|}{\begin{tabular}[c]{@{}c@{}}$N_5^\Theta(1)=43$\\ $N_5^\Theta(2)=43$\\ $N_5^\Theta(3)=42$\end{tabular}}                & \multicolumn{1}{c|}{\begin{tabular}[c]{@{}c@{}}$N_5^\Theta(1)=32$\\ $N_5^\Theta(2)=32$\\ $N_5^\Theta(3)=32$\\ $N_5^\Theta(4)=32$\end{tabular}}                 \\ \midrule
                      \multicolumn{1}{|c|}{$l_6$}                  & \multicolumn{1}{c|}{$H_6^Y(1)=112$}     & \multicolumn{1}{c|}{\begin{tabular}[c]{@{}c@{}}$H_6^Y(1)=56$\\ $H_6^Y(2)=56$\end{tabular}}       & \multicolumn{1}{c|}{\begin{tabular}[c]{@{}c@{}}$H_6^Y(1)=37$\\ $H_6^Y(2)=37$\\ $H_6^Y(3)=38$\end{tabular}}          & \multicolumn{1}{c|}{\begin{tabular}[c]{@{}c@{}}$H_6^Y(1)=28$\\ $H_6^Y(2)=28$\\ $H_6^Y(3)=28$\\ $H_6^Y(4)=28$\end{tabular}}             & \multicolumn{1}{c|}{$N_6^\Theta(1)=128$}     & \multicolumn{1}{c|}{\begin{tabular}[c]{@{}c@{}}$N_6^\Theta(1)=64$\\ $N_6^\Theta(2)=64$\end{tabular}}           & \multicolumn{1}{c|}{\begin{tabular}[c]{@{}c@{}}$N_6^\Theta(1)=43$\\ $N_6^\Theta(2)=42$\\ $N_6^\Theta(3)=43$\end{tabular}}                & \multicolumn{1}{c|}{\begin{tabular}[c]{@{}c@{}}$N_6^\Theta(1)=32$\\ $N_6^\Theta(2)=32$\\ $N_6^\Theta(3)=32$\\ $N_6^\Theta(4)=32$\end{tabular}}                 \\ \midrule
                      \multicolumn{1}{|c|}{$l_7$}                  & \multicolumn{1}{c|}{$H_7^Y(1)=56$}      & \multicolumn{1}{c|}{\begin{tabular}[c]{@{}c@{}}$H_7^Y(1)=28$\\ $H_7^Y(2)=28$\end{tabular}}       & \multicolumn{1}{c|}{\begin{tabular}[c]{@{}c@{}}$H_7^Y(1)=19$\\ $H_7^Y(2)=19$\\ $H_7^Y(3)=18$\end{tabular}}          & \multicolumn{1}{c|}{\begin{tabular}[c]{@{}c@{}}$H_7^Y(1)=14$\\ $H_7^Y(2)=14$\\ $H_7^Y(3)=14$\\ $H_7^Y(4)=14$\end{tabular}}             & \multicolumn{1}{c|}{$N_7^\Theta(1)=128$}     & \multicolumn{1}{c|}{\begin{tabular}[c]{@{}c@{}}$N_7^\Theta(1)=64$\\ $N_7^\Theta(2)=64$\end{tabular}}           & \multicolumn{1}{c|}{\begin{tabular}[c]{@{}c@{}}$N_7^\Theta(1)=42$\\ $N_7^\Theta(2)=43$\\ $N_7^\Theta(3)=43$\end{tabular}}                & \multicolumn{1}{c|}{\begin{tabular}[c]{@{}c@{}}$N_7^\Theta(1)=32$\\ $N_7^\Theta(2)=32$\\ $N_7^\Theta(3)=32$\\ $N_7^\Theta(4)=32$\end{tabular}}                 \\ \midrule
                      \multicolumn{1}{|c|}{$l_8$}                  & \multicolumn{1}{c|}{$H_8^Y(1)=56$}      & \multicolumn{1}{c|}{\begin{tabular}[c]{@{}c@{}}$H_8^Y(1)=28$\\ $H_8^Y(2)=28$\end{tabular}}       & \multicolumn{1}{c|}{\begin{tabular}[c]{@{}c@{}}$H_8^Y(1)=19$\\ $H_8^Y(2)=18$\\ $H_8^Y(3)=19$\end{tabular}}          & \multicolumn{1}{c|}{\begin{tabular}[c]{@{}c@{}}$H_8^Y(1)=14$\\ $H_8^Y(2)=14$\\ $H_8^Y(3)=14$\\ $H_8^Y(4)=14$\end{tabular}}             & \multicolumn{1}{c|}{$N_8^\Theta(1)=256$}     & \multicolumn{1}{c|}{\begin{tabular}[c]{@{}c@{}}$N_8^\Theta(1)=128$\\ $N_8^\Theta(2)=128$\end{tabular}}         & \multicolumn{1}{c|}{\begin{tabular}[c]{@{}c@{}}$N_8^\Theta(1)=86$\\ $N_8^\Theta(2)=85$\\ $N_8^\Theta(3)=85$\end{tabular}}                & \multicolumn{1}{c|}{\begin{tabular}[c]{@{}c@{}}$N_8^\Theta(1)=64$\\ $N_8^\Theta(2)=64$\\ $N_8^\Theta(3)=64$\\ $N_8^\Theta(4)=64$\end{tabular}}                 \\ \midrule
                      \multicolumn{1}{|c|}{$l_9$}                  & \multicolumn{1}{c|}{$H_9^Y(1)=56$}      & \multicolumn{1}{c|}{\begin{tabular}[c]{@{}c@{}}$H_9^Y(1)=28$\\ $H_9^Y(2)=28$\end{tabular}}       & \multicolumn{1}{c|}{\begin{tabular}[c]{@{}c@{}}$H_9^Y(1)=18$\\ $H_9^Y(2)=19$\\ $H_9^Y(3)=19$\end{tabular}}          & \multicolumn{1}{c|}{\begin{tabular}[c]{@{}c@{}}$H_9^Y(1)=14$\\ $H_9^Y(2)=14$\\ $H_9^Y(3)=14$\\ $H_9^Y(4)=14$\end{tabular}}             & \multicolumn{1}{c|}{$N_9^\Theta(1)=256$}     & \multicolumn{1}{c|}{\begin{tabular}[c]{@{}c@{}}$N_9^\Theta(1)=128$\\ $N_9^\Theta(2)=128$\end{tabular}}         & \multicolumn{1}{c|}{\begin{tabular}[c]{@{}c@{}}$N_9^\Theta(1)=85$\\ $N_9^\Theta(2)=86$\\ $N_9^\Theta(3)=85$\end{tabular}}                & \multicolumn{1}{c|}{\begin{tabular}[c]{@{}c@{}}$N_9^\Theta(1)=64$\\ $N_9^\Theta(2)=64$\\ $N_9^\Theta(3)=64$\\ $N_9^\Theta(4)=64$\end{tabular}}                 \\ \midrule
                      \multicolumn{1}{|c|}{$l_{10}$}               & \multicolumn{1}{c|}{$H_{10}^Y(1)=56$}   & \multicolumn{1}{c|}{\begin{tabular}[c]{@{}c@{}}$H_{10}^Y(1)=28$\\ $H_{10}^Y(2)=28$\end{tabular}} & \multicolumn{1}{c|}{\begin{tabular}[c]{@{}c@{}}$H_{10}^Y(1)=19$\\ $H_{10}^Y(2)=19$\\ $H_{10}^Y(3)=18$\end{tabular}} & \multicolumn{1}{c|}{\begin{tabular}[c]{@{}c@{}}$H_{10}^Y(1)=14$\\ $H_{10}^Y(2)=14$\\ $H_{10}^Y(3)=14$\\ $H_{10}^Y(4)=14$\end{tabular}} & \multicolumn{1}{c|}{$N_{10}^\Theta(1)=256$}  & \multicolumn{1}{c|}{\begin{tabular}[c]{@{}c@{}}$N_{10}^\Theta(1)=128$\\ $N_{10}^\Theta(2)=128$\end{tabular}}   & \multicolumn{1}{c|}{\begin{tabular}[c]{@{}c@{}}$N_{10}^\Theta(1)=85$\\ $N_{10}^\Theta(2)=85$\\ $N_{10}^\Theta(3)=86$\end{tabular}}       & \multicolumn{1}{c|}{\begin{tabular}[c]{@{}c@{}}$N_{10}^\Theta(1)=64$\\ $N_{10}^\Theta(2)=64$\\ $N_{10}^\Theta(3)=64$\\ $N_{10}^\Theta(4)=64$\end{tabular}}     \\ \midrule
                      \multicolumn{1}{|c|}{$l_{11}$}               & \multicolumn{1}{c|}{$H_{11}^Y(1)=28$}   & \multicolumn{1}{c|}{\begin{tabular}[c]{@{}c@{}}$H_{11}^Y(1)=14$\\ $H_{11}^Y(2)=14$\end{tabular}} & \multicolumn{1}{c|}{\begin{tabular}[c]{@{}c@{}}$H_{11}^Y(1)=9$\\ $H_{11}^Y(2)=9$\\ $H_{11}^Y(3)=10$\end{tabular}}   & \multicolumn{1}{c|}{\begin{tabular}[c]{@{}c@{}}$H_{11}^Y(1)=7$\\ $H_{11}^Y(2)=7$\\ $H_{11}^Y(3)=7$\\ $H_{11}^Y(4)=7$\end{tabular}}     & \multicolumn{1}{c|}{$N_{11}^\Theta(1)=256$}  & \multicolumn{1}{c|}{\begin{tabular}[c]{@{}c@{}}$N_{11}^\Theta(1)=128$\\ $N_{11}^\Theta(2)=128$\end{tabular}}   & \multicolumn{1}{c|}{\begin{tabular}[c]{@{}c@{}}$N_{11}^\Theta(1)=86$\\ $N_{11}^\Theta(2)=85$\\ $N_{11}^\Theta(3)=85$\end{tabular}}       & \multicolumn{1}{c|}{\begin{tabular}[c]{@{}c@{}}$N_{11}^\Theta(1)=64$\\ $N_{11}^\Theta(2)=64$\\ $N_{11}^\Theta(3)=64$\\ $N_{11}^\Theta(4)=64$\end{tabular}}     \\ \midrule
                      \multicolumn{1}{|c|}{$l_{12}$}               & \multicolumn{1}{c|}{$H_{12}^Y(1)=28$}   & \multicolumn{1}{c|}{\begin{tabular}[c]{@{}c@{}}$H_{12}^Y(1)=14$\\ $H_{12}^Y(2)=14$\end{tabular}} & \multicolumn{1}{c|}{\begin{tabular}[c]{@{}c@{}}$H_{12}^Y(1)=9$\\ $H_{12}^Y(2)=10$\\ $H_{12}^Y(3)=9$\end{tabular}}   & \multicolumn{1}{c|}{\begin{tabular}[c]{@{}c@{}}$H_{12}^Y(1)=7$\\ $H_{12}^Y(2)=7$\\ $H_{12}^Y(3)=7$\\ $H_{12}^Y(4)=7$\end{tabular}}     & \multicolumn{1}{c|}{$N_{12}^\Theta(1)=512$}  & \multicolumn{1}{c|}{\begin{tabular}[c]{@{}c@{}}$N_{12}^\Theta(1)=256$\\ $N_{12}^\Theta(2)=256$\end{tabular}}   & \multicolumn{1}{c|}{\begin{tabular}[c]{@{}c@{}}$N_{12}^\Theta(1)=171$\\ $N_{12}^\Theta(2)=171$\\ $N_{12}^\Theta(3)=170$\end{tabular}}    & \multicolumn{1}{c|}{\begin{tabular}[c]{@{}c@{}}$N_{12}^\Theta(1)=128$\\ $N_{12}^\Theta(2)=128$\\ $N_{12}^\Theta(3)=128$\\ $N_{12}^\Theta(4)=128$\end{tabular}} \\ \midrule
                      \multicolumn{1}{|c|}{$l_{13}$}               & \multicolumn{1}{c|}{$H_{13}^Y(1)=28$}   & \multicolumn{1}{c|}{\begin{tabular}[c]{@{}c@{}}$H_{13}^Y(1)=14$\\ $H_{13}^Y(2)=14$\end{tabular}} & \multicolumn{1}{c|}{\begin{tabular}[c]{@{}c@{}}$H_{13}^Y(1)=10$\\ $H_{13}^Y(2)=9$\\ $H_{13}^Y(3)=9$\end{tabular}}   & \multicolumn{1}{c|}{\begin{tabular}[c]{@{}c@{}}$H_{13}^Y(1)=7$\\ $H_{13}^Y(2)=7$\\ $H_{13}^Y(3)=7$\\ $H_{13}^Y(4)=7$\end{tabular}}     & \multicolumn{1}{c|}{$N_{13}^\Theta(1)=512$}  & \multicolumn{1}{c|}{\begin{tabular}[c]{@{}c@{}}$N_{13}^\Theta(1)=256$\\ $N_{13}^\Theta(2)=256$\end{tabular}}   & \multicolumn{1}{c|}{\begin{tabular}[c]{@{}c@{}}$N_{13}^\Theta(1)=171$\\ $N_{13}^\Theta(2)=170$\\ $N_{13}^\Theta(3)=171$\end{tabular}}    & \multicolumn{1}{c|}{\begin{tabular}[c]{@{}c@{}}$N_{13}^\Theta(1)=128$\\ $N_{13}^\Theta(2)=128$\\ $N_{13}^\Theta(3)=128$\\ $N_{13}^\Theta(4)=128$\end{tabular}} \\ \midrule
                      \multicolumn{1}{|c|}{$l_{14}$}               & \multicolumn{1}{c|}{$H_{14}^Y(1)=28$}   & \multicolumn{1}{c|}{\begin{tabular}[c]{@{}c@{}}$H_{14}^Y(1)=14$\\ $H_{14}^Y(2)=14$\end{tabular}} & \multicolumn{1}{c|}{\begin{tabular}[c]{@{}c@{}}$H_{14}^Y(1)=9$\\ $H_{14}^Y(2)=10$\\ $H_{14}^Y(3)=9$\end{tabular}}   & \multicolumn{1}{c|}{\begin{tabular}[c]{@{}c@{}}$H_{14}^Y(1)=7$\\ $H_{14}^Y(2)=7$\\ $H_{14}^Y(3)=7$\\ $H_{14}^Y(4)=7$\end{tabular}}     & \multicolumn{1}{c|}{$N_{14}^\Theta(1)=512$}  & \multicolumn{1}{c|}{\begin{tabular}[c]{@{}c@{}}$N_{14}^\Theta(1)=256$\\ $N_{14}^\Theta(2)=256$\end{tabular}}   & \multicolumn{1}{c|}{\begin{tabular}[c]{@{}c@{}}$N_{14}^\Theta(1)=170$\\ $N_{14}^\Theta(2)=171$\\ $N_{14}^\Theta(3)=171$\end{tabular}}    & \multicolumn{1}{c|}{\begin{tabular}[c]{@{}c@{}}$N_{14}^\Theta(1)=128$\\ $N_{14}^\Theta(2)=128$\\ $N_{14}^\Theta(3)=128$\\ $N_{14}^\Theta(4)=128$\end{tabular}} \\ \midrule
                      \multicolumn{1}{|c|}{$l_{15}$}               & \multicolumn{1}{c|}{$H_{15}^Y(1)=14$}   & \multicolumn{1}{c|}{\begin{tabular}[c]{@{}c@{}}$H_{15}^Y(1)=7$\\ $H_{15}^Y(2)=7$\end{tabular}}   & \multicolumn{1}{c|}{\begin{tabular}[c]{@{}c@{}}$H_{15}^Y(1)=5$\\ $H_{15}^Y(2)=5$\\ $H_{15}^Y(3)=4$\end{tabular}}    & \multicolumn{1}{c|}{\begin{tabular}[c]{@{}c@{}}$H_{15}^Y(1)=4$\\ $H_{15}^Y(2)=4$\\ $H_{15}^Y(3)=3$\\ $H_{15}^Y(4)=3$\end{tabular}}     & \multicolumn{1}{c|}{$N_{15}^\Theta(1)=512$}  & \multicolumn{1}{c|}{\begin{tabular}[c]{@{}c@{}}$N_{15}^\Theta(1)=256$\\ $N_{15}^\Theta(2)=256$\end{tabular}}   & \multicolumn{1}{c|}{\begin{tabular}[c]{@{}c@{}}$N_{15}^\Theta(1)=171$\\ $N_{15}^\Theta(2)=171$\\ $N_{15}^\Theta(3)=170$\end{tabular}}    & \multicolumn{1}{c|}{\begin{tabular}[c]{@{}c@{}}$N_{15}^\Theta(1)=128$\\ $N_{15}^\Theta(2)=128$\\ $N_{15}^\Theta(3)=128$\\ $N_{15}^\Theta(4)=128$\end{tabular}} \\ \midrule
                      \multicolumn{1}{|c|}{$l_{16}$}               & \multicolumn{1}{c|}{$H_{16}^Y(1)=14$}   & \multicolumn{1}{c|}{\begin{tabular}[c]{@{}c@{}}$H_{16}^Y(1)=7$\\ $H_{16}^Y(2)=7$\end{tabular}}   & \multicolumn{1}{c|}{\begin{tabular}[c]{@{}c@{}}$H_{16}^Y(1)=5$\\ $H_{16}^Y(2)=4$\\ $H_{16}^Y(3)=5$\end{tabular}}    & \multicolumn{1}{c|}{\begin{tabular}[c]{@{}c@{}}$H_{16}^Y(1)=3$\\ $H_{16}^Y(2)=3$\\ $H_{16}^Y(3)=4$\\ $H_{16}^Y(4)=4$\end{tabular}}     & \multicolumn{1}{c|}{$N_{16}^\Theta(1)=512$}  & \multicolumn{1}{c|}{\begin{tabular}[c]{@{}c@{}}$N_{16}^\Theta(1)=256$\\ $N_{16}^\Theta(2)=256$\end{tabular}}   & \multicolumn{1}{c|}{\begin{tabular}[c]{@{}c@{}}$N_{16}^\Theta(1)=171$\\ $N_{16}^\Theta(2)=170$\\ $N_{16}^\Theta(3)=171$\end{tabular}}    & \multicolumn{1}{c|}{\begin{tabular}[c]{@{}c@{}}$N_{16}^\Theta(1)=128$\\ $N_{16}^\Theta(2)=128$\\ $N_{16}^\Theta(3)=128$\\ $N_{16}^\Theta(4)=128$\end{tabular}} \\ \midrule
                      \multicolumn{1}{|c|}{$l_{17}$}               & \multicolumn{1}{c|}{$H_{17}^Y(1)=14$}   & \multicolumn{1}{c|}{\begin{tabular}[c]{@{}c@{}}$H_{17}^Y(1)=7$\\ $H_{17}^Y(2)=7$\end{tabular}}   & \multicolumn{1}{c|}{\begin{tabular}[c]{@{}c@{}}$H_{17}^Y(1)=4$\\ $H_{17}^Y(2)=5$\\ $H_{17}^Y(3)=5$\end{tabular}}    & \multicolumn{1}{c|}{\begin{tabular}[c]{@{}c@{}}$H_{17}^Y(1)=3$\\ $H_{17}^Y(2)=4$\\ $H_{17}^Y(3)=3$\\ $H_{17}^Y(4)=4$\end{tabular}}     & \multicolumn{1}{c|}{$N_{17}^\Theta(1)=512$}  & \multicolumn{1}{c|}{\begin{tabular}[c]{@{}c@{}}$N_{17}^\Theta(1)=256$\\ $N_{17}^\Theta(2)=256$\end{tabular}}   & \multicolumn{1}{c|}{\begin{tabular}[c]{@{}c@{}}$N_{17}^\Theta(1)=170$\\ $N_{17}^\Theta(2)=171$\\ $N_{17}^\Theta(3)=171$\end{tabular}}    & \multicolumn{1}{c|}{\begin{tabular}[c]{@{}c@{}}$N_{17}^\Theta(1)=128$\\ $N_{17}^\Theta(2)=128$\\ $N_{17}^\Theta(3)=128$\\ $N_{17}^\Theta(4)=128$\end{tabular}} \\ \midrule
                      \multicolumn{1}{|c|}{$l_{18}$}               & \multicolumn{1}{c|}{$H_{18}^Y(1)=14$}   & \multicolumn{1}{c|}{\begin{tabular}[c]{@{}c@{}}$H_{18}^Y(1)=7$\\ $H_{18}^Y(2)=7$\end{tabular}}   & \multicolumn{1}{c|}{\begin{tabular}[c]{@{}c@{}}$H_{18}^Y(1)=4$\\ $H_{18}^Y(2)=5$\\ $H_{18}^Y(3)=5$\end{tabular}}    & \multicolumn{1}{c|}{\begin{tabular}[c]{@{}c@{}}$H_{18}^Y(1)=4$\\ $H_{18}^Y(2)=3$\\ $H_{18}^Y(3)=4$\\ $H_{18}^Y(4)=3$\end{tabular}}     & \multicolumn{1}{c|}{$N_{18}^\Theta(1)=512$}  & \multicolumn{1}{c|}{\begin{tabular}[c]{@{}c@{}}$N_{18}^\Theta(1)=256$\\ $N_{18}^\Theta(2)=256$\end{tabular}}   & \multicolumn{1}{c|}{\begin{tabular}[c]{@{}c@{}}$N_{18}^\Theta(1)=171$\\ $N_{18}^\Theta(2)=171$\\ $N_{18}^\Theta(3)=170$\end{tabular}}    & \multicolumn{1}{c|}{\begin{tabular}[c]{@{}c@{}}$N_{18}^\Theta(1)=128$\\ $N_{18}^\Theta(2)=128$\\ $N_{18}^\Theta(3)=128$\\ $N_{18}^\Theta(4)=128$\end{tabular}} \\ \midrule
                      \multicolumn{1}{|c|}{$l_{19}$}               & \multicolumn{1}{c|}{$H_{19}^Y(1)=7$}    & \multicolumn{1}{c|}{\begin{tabular}[c]{@{}c@{}}$H_{19}^Y(1)=4$\\ $H_{19}^Y(2)=3$\end{tabular}}   & \multicolumn{1}{c|}{\begin{tabular}[c]{@{}c@{}}$H_{19}^Y(1)=2$\\ $H_{19}^Y(2)=2$\\ $H_{19}^Y(3)=3$\end{tabular}}    & \multicolumn{1}{c|}{\begin{tabular}[c]{@{}c@{}}$H_{19}^Y(1)=2$\\ $H_{19}^Y(2)=2$\\ $H_{19}^Y(3)=2$\\ $H_{19}^Y(4)=1$\end{tabular}}     & \multicolumn{1}{c|}{$N_{19}^\Theta(1)=512$}  & \multicolumn{1}{c|}{\begin{tabular}[c]{@{}c@{}}$N_{19}^\Theta(1)=256$\\ $N_{19}^\Theta(2)=256$\end{tabular}}   & \multicolumn{1}{c|}{\begin{tabular}[c]{@{}c@{}}$N_{19}^\Theta(1)=171$\\ $N_{19}^\Theta(2)=170$\\ $N_{19}^\Theta(3)=171$\end{tabular}}    & \multicolumn{1}{c|}{\begin{tabular}[c]{@{}c@{}}$N_{19}^\Theta(1)=128$\\ $N_{19}^\Theta(2)=128$\\ $N_{19}^\Theta(3)=128$\\ $N_{19}^\Theta(4)=128$\end{tabular}} \\ \midrule
                      \multicolumn{1}{|c|}{$l_{20}$}               & \multicolumn{1}{c|}{$H_{20}^Y(1)=1$} & \multicolumn{1}{c|}{\begin{tabular}[c]{@{}c@{}}$H_{20}^Y(1)=1$\\ $H_{20}^Y(2)=0$\end{tabular}}   & \multicolumn{1}{c|}{\begin{tabular}[c]{@{}c@{}}$H_{20}^Y(1)=1$\\ $H_{20}^Y(2)=0$\\ $H_{20}^Y(3)=0$\end{tabular}}    & \multicolumn{1}{c|}{\begin{tabular}[c]{@{}c@{}}$H_{20}^Y(1)=1$\\ $H_{20}^Y(2)=0$\\ $H_{20}^Y(3)=0$\\ $H_{20}^Y(4)=0$\end{tabular}}  & \multicolumn{1}{c|}{$N_{20}^\Theta(1)=4096$} & \multicolumn{1}{c|}{\begin{tabular}[c]{@{}c@{}}$N_{20}^\Theta(1)=2048$\\ $N_{20}^\Theta(2)=2048$\end{tabular}} & \multicolumn{1}{c|}{\begin{tabular}[c]{@{}c@{}}$N_{20}^\Theta(1)=1365$\\ $N_{20}^\Theta(2)=1365$\\ $N_{20}^\Theta(3)=1366$\end{tabular}} & \multicolumn{1}{c|}{\begin{tabular}[c]{@{}c@{}}$N_{20}^\Theta(1)=512$\\ $N_{20}^\Theta(2)=512$\\ $N_{20}^\Theta(3)=512$\\ $N_{20}^\Theta(4)=512$\end{tabular}} \\ \midrule
                      \multicolumn{1}{|c|}{$l_{21}$}               & \multicolumn{1}{c|}{$H_{21}^Y(1)=1$} & \multicolumn{1}{c|}{\begin{tabular}[c]{@{}c@{}}$H_{21}^Y(1)=0$\\ $H_{21}^Y(2)=1$\end{tabular}}   & \multicolumn{1}{c|}{\begin{tabular}[c]{@{}c@{}}$H_{21}^Y(1)=0$\\ $H_{21}^Y(2)=1$\\ $H_{21}^Y(3)=0$\end{tabular}}    & \multicolumn{1}{c|}{\begin{tabular}[c]{@{}c@{}}$H_{21}^Y(1)=0$\\ $H_{21}^Y(2)=1$\\ $H_{21}^Y(3)=0$\\ $H_{21}^Y(4)=0$\end{tabular}}  & \multicolumn{1}{c|}{$N_{21}^\Theta(1)=4096$} & \multicolumn{1}{c|}{\begin{tabular}[c]{@{}c@{}}$N_{21}^\Theta(1)=2048$\\ $N_{21}^\Theta(2)=2048$\end{tabular}} & \multicolumn{1}{c|}{\begin{tabular}[c]{@{}c@{}}$N_{21}^\Theta(1)=1366$\\ $N_{21}^\Theta(2)=1365$\\ $N_{21}^\Theta(3)=1365$\end{tabular}} & \multicolumn{1}{c|}{\begin{tabular}[c]{@{}c@{}}$N_{21}^\Theta(1)=512$\\ $N_{21}^\Theta(2)=512$\\ $N_{21}^\Theta(3)=512$\\ $N_{21}^\Theta(4)=512$\end{tabular}} \\ \midrule
                      \multicolumn{1}{|c|}{$l_{22}$}               & \multicolumn{1}{c|}{$H_{22}^Y(1)=1$} & \multicolumn{1}{c|}{\begin{tabular}[c]{@{}c@{}}$H_{22}^Y(1)=1$\\ $H_{22}^Y(2)=0$\end{tabular}}   & \multicolumn{1}{c|}{\begin{tabular}[c]{@{}c@{}}$H_{22}^Y(1)=0$\\ $H_{22}^Y(2)=0$\\ $H_{22}^Y(3)=1$\end{tabular}}    & \multicolumn{1}{c|}{\begin{tabular}[c]{@{}c@{}}$H_{22}^Y(1)=0$\\ $H_{22}^Y(2)=0$\\ $H_{22}^Y(3)=1$\\ $H_{22}^Y(4)=0$\end{tabular}}  & \multicolumn{1}{c|}{$N_{22}^\Theta(1)=1000$} & \multicolumn{1}{c|}{\begin{tabular}[c]{@{}c@{}}$N_{22}^\Theta(1)=500$\\ $N_{22}^\Theta(2)=500$\end{tabular}}   & \multicolumn{1}{c|}{\begin{tabular}[c]{@{}c@{}}$N_{22}^\Theta(1)=333$\\ $N_{22}^\Theta(2)=334$\\ $N_{22}^\Theta(3)=333$\end{tabular}}    & \multicolumn{1}{c|}{\begin{tabular}[c]{@{}c@{}}$N_{22}^\Theta(1)=250$\\ $N_{22}^\Theta(2)=250$\\ $N_{22}^\Theta(3)=250$\\ $N_{22}^\Theta(4)=250$\end{tabular}} \\ \midrule
                      \multicolumn{1}{|c|}{$l_{23}$}                                    & \multicolumn{1}{c|}{$H_{23}^Y(1)=1$}                      & \multicolumn{1}{c|}{\begin{tabular}[c]{@{}c@{}}$H_{23}^Y(1)=0$\\ $H_{23}^Y(2)=1$\end{tabular}}                        & \multicolumn{1}{c|}{\begin{tabular}[c]{@{}c@{}}$H_{23}^Y(1)=1$\\ $H_{23}^Y(2)=0$\\ $H_{23}^Y(3)=0$\end{tabular}}                         & \multicolumn{1}{c|}{\begin{tabular}[c]{@{}c@{}}$H_{23}^Y(1)=0$\\ $H_{23}^Y(2)=0$\\ $H_{23}^Y(3)=0$\\ $H_{23}^Y(4)=1$\end{tabular}}                       & \multicolumn{1}{c|}{N.A.}                                         & \multicolumn{1}{c|}{N.A.}                                                                                                           & \multicolumn{1}{c|}{N.A.}                                                                                                                                     & \multicolumn{1}{c|}{N.A.}                                                                                                                                                           \\ \bottomrule
                      
			\end{tabular}}}
			{%
				\caption{VGG 16 partitions specification: Part 1}%
				\label{tab:VGG16_partitions_specification1}
			}
		\end{floatrow}
	\end{figure*}

\begin{table*}[!t]
	\centering
	\resizebox{\columnwidth}{!}{%
	\begin{tabular}{|ccc|c|c|c|c|c|c|c|c|c|c|c|c|c|c|c|c|c|c|c|c|c|c|c|}
		\hline
		\multicolumn{3}{|c|}{layer}                                                                                                                                                                                           & $l_1$    & $l_2$    & $l_3$    & $l_4$    & $l_5$    & $l_6$    & $l_7$    & $l_8$    & $l_9$    & $l_{10}$ & $l_{11}$ & $l_{12}$ & $l_{13}$ & $l_{14}$ & $l_{15}$ & $l_{16}$ & $l_{17}$ & $l_{18}$ & $l_{19}$ & $l_{20}$ & $l_{21}$ & $l_{22}$ & $l_{23}$ \\ \hline
		\multicolumn{1}{|c|}{\multirow{4}{*}{\begin{tabular}[c]{@{}c@{}}Sequencial \\ Partitioning\end{tabular}}} & \multicolumn{1}{c|}{\multirow{4}{*}{\begin{tabular}[c]{@{}c@{}}number of \\ partitions\end{tabular}}} & 1 & $p_1$    & $p_1$    & $p_1$    & $p_1$    & $p_1$    & $p_1$    & $p_1$    & $p_1$    & $p_1$    & $p_1$    & $p_1$    & $p_1$    & $p_1$    & $p_1$    & $p_1$    & $p_1$    & $p_1$    & $p_1$    & $p_1$    & $p_1$    & $p_1$    & $p_1$    & $p_1$    \\ \cline{3-26} 
		\multicolumn{1}{|c|}{}                                                                                    & \multicolumn{1}{c|}{}                                                                                 & 2 & $p_1$    & $p_1$    & $p_1$    & $p_1$    & $p_1$    & $p_1$    & $p_1$    & $p_1$    & $p_1$    & $p_1$    & $p_1$    & $p_1$    & $p_2$    & $p_2$    & $p_2$    & $p_2$    & $p_2$    & $p_2$    & $p_2$    & $p_2$    & $p_2$    & $p_2$    & $p_2$    \\ \cline{3-26} 
		\multicolumn{1}{|c|}{}                                                                                    & \multicolumn{1}{c|}{}                                                                                 & 3 & $p_1$    & $p_1$    & $p_1$    & $p_1$    & $p_1$    & $p_1$    & $p_1$    & $p_1$    & $p_2$    & $p_2$    & $p_2$    & $p_2$    & $p_2$    & $p_2$    & $p_2$    & $p_3$    & $p_3$    & $p_3$    & $p_3$    & $p_3$    & $p_3$    & $p_3$    & $p_3$    \\ \cline{3-26} 
		\multicolumn{1}{|c|}{}                                                                                    & \multicolumn{1}{c|}{}                                                                                 & 4 & $p_1$    & $p_1$    & $p_1$    & $p_1$    & $p_1$    & $p_1$    & $p_2$    & $p_2$    & $p_2$    & $p_2$    & $p_3$    & $p_3$    & $p_3$    & $p_3$    & $p_3$    & $p_4$    & $p_4$    & $p_4$    & $p_4$    & $p_4$    & $p_4$    & $p_4$    & $p_4$    \\ \hline
		\multicolumn{1}{|c|}{\multirow{4}{*}{\begin{tabular}[c]{@{}c@{}}Vertical \\ Partitioning\end{tabular}}}   & \multicolumn{1}{c|}{\multirow{4}{*}{\begin{tabular}[c]{@{}c@{}}number of \\ partitions\end{tabular}}} & 1 & $p_{11}$ & $p_{11}$ & $p_{11}$ & $p_{11}$ & $p_{11}$ & $p_{11}$ & $p_{11}$ & $p_{11}$ & $p_{11}$ & $p_{11}$ & $p_{11}$ & $p_{11}$ & $p_{11}$ & $p_{11}$ & $p_{11}$ & $p_{11}$ & $p_{11}$ & $p_{11}$ & $p_{11}$ & $p_{11}$ & $p_{11}$ & $p_{11}$ & $p_{11}$ \\ \cline{3-26} 
		\multicolumn{1}{|c|}{}                                                                                    & \multicolumn{1}{c|}{}                                                                                 & 2 & $p_{11}$ & $p_{11}$ & $p_{11}$ & $p_{11}$ & $p_{11}$ & $p_{11}$ & $p_{11}$ & $p_{11}$ & $p_{21}$ & $p_{21}$ & $p_{21}$ & $p_{21}$ & $p_{21}$ & $p_{12}$ & $p_{12}$ & $p_{12}$ & $p_{22}$ & $p_{22}$ & $p_{22}$ & $p_{22}$ & $p_{22}$ & $p_{22}$ & $p_{22}$ \\ \cline{3-26} 
		\multicolumn{1}{|c|}{}                                                                                    & \multicolumn{1}{c|}{}                                                                                 & 3 & $p_{11}$ & $p_{11}$ & $p_{11}$ & $p_{11}$ & $p_{11}$ & $p_{21}$ & $p_{21}$ & $p_{21}$ & $p_{31}$ & $p_{31}$ & $p_{31}$ & $p_{12}$ & $p_{12}$ & $p_{12}$ & $p_{12}$ & $p_{22}$ & $p_{22}$ & $p_{22}$ & $p_{22}$ & $p_{22}$ & $p_{32}$ & $p_{32}$ & $p_{32}$ \\ \cline{3-26} 
		\multicolumn{1}{|c|}{}                                                                                    & \multicolumn{1}{c|}{}                                                                                 & 4 & $p_{11}$ & $p_{11}$ & $p_{11}$ & $p_{11}$ & $p_{11}$ & $p_{21}$ & $p_{21}$ & $p_{21}$ & $p_{31}$ & $p_{41}$ & $p_{41}$ & $p_{41}$ & $p_{41}$ & $p_{12}$ & $p_{22}$ & $p_{22}$ & $p_{32}$ & $p_{32}$ & $p_{32}$ & $p_{32}$ & $p_{32}$ & $p_{32}$ & $p_{32}$ \\ \hline
	\end{tabular}%
  }
\caption{VGG 16 partitions specification: Part 2}
\label{tab:VGG16_partitions_specification2}
\end{table*}

Using the Data Partitioning Strategy, the output data tensor of every layer is partitioned into one, two, three, or four sub-tensors, depending on the number of partitions that are consider. The partitioning is done in such a way that the difference among the heights of the sub-tensors is minimized. When VGG 16, shown in Figure~\ref{fig:VGG16_network_structure}, is distributed and implemented by using the Data Partitioning Strategy, the partitions specification is shown in Table~\ref{tab:VGG16_partitions_specification1} - see Columns 2-5. Each row shows one layer of VGG 16 and each column shows the number of partitions that are considered. For example, Row 5 in Column 4 shows that when layer $l_4$ is distributed onto three partitions, its output data tensor is partitioned into three sub-tensors where the heights of the sub-tensors are $H_4^Y(1)=38$, $H_4^Y(2)=37$, and $H_4^Y(3)=37$ for the first, second, and third sub-tensor, respectively.

Using the Horizontal Partitioning Strategy, the weights/neurons of every layer are partitioned into one, two, three, or four groups, depending on the number of partitions that are considered. The partitioning is done in such a way that the difference among the number of neurons in the groups is minimized. When VGG 16, shown in Figure~\ref{fig:VGG16_network_structure}, is distributed and implemented by using the Horizontal Partitioning Strategy, the partitions specification is shown in Table~\ref{tab:VGG16_partitions_specification1} - see Columns 6-9. Each row shows one layer of VGG 16 and each column shows the number of partitions that are considered. For example, Row 9 in Column 9 shows that when layer $l_8$ is distributed onto four partitions, its neurons are partitioned into four groups where the number of neurons are $N_8^\Theta(1)=64$, $N_8^\Theta(2)=64$, $N_8^\Theta(3)=64$, and $N_8^\Theta(4)=64$ in the first, second, third, and fourth group, respectively.

Using the Sequential Partitioning Strategy, the layers of a CNN model are partitioned into one, two, three, or four partitions, depending on the number of partitions that are considered. The partitioning is done in such a way that each partition consists of consecutive layers and the difference among the energy consumption of the partitions is minimized. Such partitioning is obtained by using a genetic algorithm. When VGG 16, shown in Figure~\ref{fig:VGG16_network_structure}, is distributed and implemented by using the Sequential Partitioning Strategy, the partitions specification is shown in Table~\ref{tab:VGG16_partitions_specification2} - see Rows 2-5. Each column shows one layer of VGG 16 and each row shows the number of partitions that are considered. For example, Row 3 and Column 5 show that when VGG 16 is distributed onto two partitions, layer $l_2$ is part of partition $p_1$.

Using the Vertical Partitioning Strategy, the layers of a CNN model are partitioned into one, two, three, or four partitions, depending on the number of partitions that are considered, where each partition may consist of several sub-partitions. The partitioning is done in such a way that every partition may contain non-consecutive layers but the layers in each of its sub-partitions are consecutive, and the difference among the energy consumption of the partitions is minimized. Such partitioning is obtained by using a genetic algorithm. When VGG 16, shown in Figure~\ref{fig:VGG16_network_structure}, is distributed and implemented by using the Vertical Partitioning Strategy, the partitions specification is shown in Table~\ref{tab:VGG16_partitions_specification2} - see Rows 6-9. Each column shows one layer of VGG 16 and each row shows the number of partitions that are considered. For example, Row 9 shows that when VGG 16 is distributed onto four partitions then partition $p_{1}$ consists of two sub-partitions $p_{11}$ and $p_{12}$. Similarly, $p_{2}$ and $p_{3}$ have two sub-partitions each. Column 9 specifies that layer $l_6$ is part of sub-partition $p_{21}$ of partition $p_{2}$.

Similar to the aforementioned setup for VGG 16, by utilizing the four partitioning strategies, the Emotion\_fer model is partitioned, mapped, and executed on a homogeneous distributed system at the edge which consists of four NVIDIA Jetson TX2 embedded platforms. For every distributed system implementation, obtained by applying the four different partitioning strategies on VGG 16 and Emotion\_fer, the energy consumption per edge device is directly measured on the distributed system at the edge, and the average value over 500 CNN inference executions is taken. In addition, the energy consumption per edge device is estimated by using our analytical models. To do this, every layer in VGG 16 and Emotion\_fer is profiled on the NVIDIA Jetson TX2 platform to obtain the power consumption and execution time data needed to calibrate our energy consumption analytical models introduced in Section~\ref{sec:model}. After this model calibration, the energy consumption per edge device is calculated by using these analytical models. 

\subsection{Accuracy of the energy models}
\label{sec:accuracy}
In this section, we present the experimental results, we have obtained for VGG 16 and Emotion\_fer, in order to confirm the accuracy of our energy consumption analytical models, introduced in Section~\ref{sec:model}, for the four partitioning strategies that enable distributed CNN model inference at the edge. We compare the energy consumption per edge device $E_j$, estimated by using our analytical models, with the corresponding numbers, obtained by direct measurements on the distributed system implementations using the experimental setup, described in Section~\ref{sec:Experimental_setup}. The results are shown in Table~\ref{table:error}. 

\begin{table*}[!t]
	\centering
	\caption{Accuracy evaluation of our energy consumption analytical models}
	\label{table:error}
	\begin{adjustbox}{center, width=0.7\columnwidth}
	\begin{tabular}{|c|c|ccc|ccc|}
		\hline
		\multirow{2}{*}{\begin{tabular}[c]{@{}c@{}}Partitioning \\ Strategy\end{tabular}}               & \multirow{2}{*}{\begin{tabular}[c]{@{}c@{}}Number of \\ partitions\end{tabular}} & \multicolumn{3}{c|}{VGG 16}                                                                                                                                                                                                                                                                      & \multicolumn{3}{c|}{Emotion\_fer}                                                                                                                                                                                                                                                                \\ \cline{3-8} 
		&                                                                                  & \multicolumn{1}{c|}{\begin{tabular}[c]{@{}c@{}}energy consumption \\ by analytical models\\  (J/img)\end{tabular}} & \multicolumn{1}{c|}{\begin{tabular}[c]{@{}c@{}}energy consumption \\ by measurement \\ (J/img)\end{tabular}} & \begin{tabular}[c]{@{}c@{}}energy error\\  (\%)\end{tabular} & \multicolumn{1}{c|}{\begin{tabular}[c]{@{}c@{}}energy consumption \\ by analytical models\\  (J/img)\end{tabular}} & \multicolumn{1}{c|}{\begin{tabular}[c]{@{}c@{}}energy consumption \\ by measurement \\ (J/img)\end{tabular}} & \begin{tabular}[c]{@{}c@{}}energy error \\ (\%)\end{tabular} \\ \hline
		\multirow{4}{*}{\begin{tabular}[c]{@{}c@{}}Data \\ Partitioning \\ Strategy\end{tabular}}       & 1                                                                                & \multicolumn{1}{c|}{0.318}                                                                                         & \multicolumn{1}{c|}{0.321}                                                                                   & -0.9                                                         & \multicolumn{1}{c|}{0.0175}                                                                                        & \multicolumn{1}{c|}{0.0172}                                                                                  & 1.7                                                          \\ \cline{2-8} 
		& 2                                                                                & \multicolumn{1}{c|}{0.221}                                                                                         & \multicolumn{1}{c|}{0.225}                                                                                   & -1.8                                                         & \multicolumn{1}{c|}{0.0161}                                                                                        & \multicolumn{1}{c|}{0.0166}                                                                                  & -3.0                                                         \\ \cline{2-8} 
		& 3                                                                                & \multicolumn{1}{c|}{0.174}                                                                                         & \multicolumn{1}{c|}{0.162}                                                                                   & 7.4                                                          & \multicolumn{1}{c|}{0.0161}                                                                                        & \multicolumn{1}{c|}{0.0169}                                                                                  & -4.7                                                         \\ \cline{2-8} 
		& 4                                                                                & \multicolumn{1}{c|}{0.123}                                                                                         & \multicolumn{1}{c|}{0.129}                                                                                   & -4.7                                                         & \multicolumn{1}{c|}{0.0157}                                                                                        & \multicolumn{1}{c|}{0.0170}                                                                                  & -7.6                                                         \\ \hline
		\multirow{4}{*}{\begin{tabular}[c]{@{}c@{}}Horizontal \\ Partitioning \\ Strategy\end{tabular}} & 1                                                                                & \multicolumn{1}{c|}{0.318}                                                                                         & \multicolumn{1}{c|}{0.321}                                                                                   & -0.9                                                         & \multicolumn{1}{c|}{0.0175}                                                                                        & \multicolumn{1}{c|}{0.0172}                                                                                  & 1.7                                                          \\ \cline{2-8} 
		& 2                                                                                & \multicolumn{1}{c|}{0.249}                                                                                         & \multicolumn{1}{c|}{0.241}                                                                                   & 3.3                                                          & \multicolumn{1}{c|}{0.0298}                                                                                        & \multicolumn{1}{c|}{0.0287}                                                                                  & 3.7                                                          \\ \cline{2-8} 
		& 3                                                                                & \multicolumn{1}{c|}{0.165}                                                                                         & \multicolumn{1}{c|}{0.177}                                                                                   & -6.8                                                         & \multicolumn{1}{c|}{0.0754}                                                                                        & \multicolumn{1}{c|}{0.0776}                                                                                  & -2.8                                                         \\ \cline{2-8} 
		& 4                                                                                & \multicolumn{1}{c|}{0.147}                                                                                         & \multicolumn{1}{c|}{0.140}                                                                                   & 5.0                                                          & \multicolumn{1}{c|}{0.0516}                                                                                        & \multicolumn{1}{c|}{0.0530}                                                                                  & -2.6                                                         \\ \hline
		\multirow{4}{*}{\begin{tabular}[c]{@{}c@{}}Sequential \\ Partitioning \\ Strategy\end{tabular}} & 1                                                                                & \multicolumn{1}{c|}{0.318}                                                                                         & \multicolumn{1}{c|}{0.321}                                                                                   & -0.9                                                         & \multicolumn{1}{c|}{0.0175}                                                                                        & \multicolumn{1}{c|}{0.0172}                                                                                  & 1.7                                                          \\ \cline{2-8} 
		& 2                                                                                & \multicolumn{1}{c|}{0.187}                                                                                         & \multicolumn{1}{c|}{0.192}                                                                                   & -2.6                                                         & \multicolumn{1}{c|}{0.0127}                                                                                        & \multicolumn{1}{c|}{0.0121}                                                                                  & 5.0                                                          \\ \cline{2-8} 
		& 3                                                                                & \multicolumn{1}{c|}{0.147}                                                                                         & \multicolumn{1}{c|}{0.144}                                                                                   & 2.1                                                          & \multicolumn{1}{c|}{0.0081}                                                                                        & \multicolumn{1}{c|}{0.0087}                                                                                  & -6.9                                                         \\ \cline{2-8} 
		& 4                                                                                & \multicolumn{1}{c|}{0.102}                                                                                         & \multicolumn{1}{c|}{0.108}                                                                                   & -5.6                                                         & \multicolumn{1}{c|}{0.0071}                                                                                        & \multicolumn{1}{c|}{0.0068}                                                                                  & 4.4                                                          \\ \hline
		\multirow{4}{*}{\begin{tabular}[c]{@{}c@{}}Vertical\\  Partitioning \\ Strategy\end{tabular}}   & 1                                                                                & \multicolumn{1}{c|}{0.318}                                                                                         & \multicolumn{1}{c|}{0.321}                                                                                   & -0.9                                                         & \multicolumn{1}{c|}{0.0175}                                                                                        & \multicolumn{1}{c|}{0.0172}                                                                                  & 1.7                                                          \\ \cline{2-8} 
		& 2                                                                                & \multicolumn{1}{c|}{0.174}                                                                                         & \multicolumn{1}{c|}{0.178}                                                                                   & -2.2                                                         & \multicolumn{1}{c|}{0.0125}                                                                                        & \multicolumn{1}{c|}{0.0119}                                                                                  & 5.0                                                          \\ \cline{2-8} 
		& 3                                                                                & \multicolumn{1}{c|}{0.122}                                                                                         & \multicolumn{1}{c|}{0.127}                                                                                   & -3.9                                                         & \multicolumn{1}{c|}{0.0083}                                                                                        & \multicolumn{1}{c|}{0.0084}                                                                                  & -1.2                                                         \\ \cline{2-8} 
		& 4                                                                                & \multicolumn{1}{c|}{0.093}                                                                                         & \multicolumn{1}{c|}{0.087}                                                                                   & 6.9                                                          & \multicolumn{1}{c|}{0.0069}                                                                                        & \multicolumn{1}{c|}{0.0065}                                                                                  & 6.2                                                          \\ \hline
	\end{tabular}
\end{adjustbox}
\end{table*}

In Row 1, we list the two experimental CNN models (VGG 16 and Emotion\_fer), mentioned in Section~\ref{sec:Experimental_setup}. Column 1 shows the partitioning strategies and Column 2 indicates the number of partitions that are considered. Column 3 and 4 show the estimated and measured maximum energy consumption per edge device (in Joules per image) for VGG 16. Column 5 shows the difference (i.e., the error in \%) between the estimated and measured energy consumption. The same information is shown in Column 6, 7, and 8 for Emotion\_fer. In Column 5 and 8, a negative error value means that the maximum energy consumption per edge device is underestimated and a positive value means that the maximum energy consumption per edge device is overestimated by our analytical models. We can see in Table~\ref{table:error} that the error rate for the maximum energy consumption per edge device is below 8\%. This fact confirms that our analytical models are sufficiently accurate to relatively compare different partitioning strategies, in terms of energy consumption per edge device, for CNN model inference at the edge. 
\section{Large-scale Investigation and Discussion}
\label{sec:investigation}
In this section, we utilize our accurate energy consumption analytical models, introduced in Section~\ref{sec:model}, in a large-scale experiment including the nine representative CNN models from the ONNX model zoo library, introduced in Section~\ref{sec:CNNs}. In this experiment, we investigate for every representative CNN model, inferred on distributed devices at the edge, which partitioning strategy has the highest potential to decrease the maximum energy consumption per edge device. First, in Section~\ref{sec:large_setup}, we describe the experimental setup for our large-scale investigation. Then, in Section~\ref{sec:investigation_results}, we present and analyse the results, we have obtained for each of the nine representative CNN models. Finally, in Section~\ref{sec:Discussion}, we summarize and discuss our general findings on the potential of the four partitioning strategies to decrease the maximum energy consumption per edge device. 

\subsection{Experimental Setup for Large-scale Investigation}
\label{sec:large_setup}
We conduct our experiment using the nine CNN models listed in Table~\ref{tab:CNNs}. First, for every CNN model, we construct a model network structure and partitions specification tables in the same way as described in Section~\ref{sec:Experimental_setup} for the VGG 16 model -- see Figure~\ref{fig:VGG16_network_structure}, Table~\ref{tab:VGG16_partitions_specification1}, and Table~\ref{tab:VGG16_partitions_specification2} as examples for VGG 16. In this large-scale experiment, for each CNN model, the partitions specification tables are constructed considering up to ten edge devices/partitions for each partitioning strategy evaluation. 

Second, to properly calibrate our energy consumption analytical models introduced in Section~\ref{sec:model}, we have to set accurate values of all device-dependent parameters used in the equations described in Section~\ref{sec:model}. For every layer $l_i$ in a CNN model, these parameters are: $t_i$, $P_{i} \left ( t \right )$, $E_{in,comm}^{i}\left(H^Y_i \times W^Y_i  \times C^Y_i \right)$, $E_{ex,comm}^{i}\left(H^Y_i \times W^Y_i  \times C^Y_i \right)$, $E_{ex,send}^{b}\left (  H^Y_b  \times W^Y_b \times C^Y_b  \right )$ and $E_{ex,receive}^{a}\left (  H^X_a  \times W^X_a \times C^X_a  \right )$. We obtain accurate values for these model parameters by executing/profiling every CNN model, layer by layer, and taking real measurements on the NVIDIA Jetson TX2 platform. The features of this platform are briefly described in Section~\ref{sec:Distributed_hardware_platform} and we assume that this platform is used in every device belonging to a distributed system configuration with multiple edge devices. The energy consumption models calibration, discussed above, is a time consuming effort but it is a one-time effort. After this initial effort, the calibrated energy models can be used to estimate, very fast and accurately, the per-device energy consumption of a CNN inferred on many different distributed system configurations at the edge without actual CNN deployment on these configurations. It is worth noting that actual CNN deployment on a distributed system configuration is a very time consuming process that hinders large-scale investigations and experiments.  

Finally, for every CNN model, we use the information in the aforementioned model network structure and partitions specification tables together with the calibrated energy consumption analytical models to compute the energy consumption per edge device when the CNN model is partitioned using the four partitioning strategies and inferred on the distributed systems at the edge specified in the partitions tables. The number of partitions mentioned in these tables is equal to the number of edge devices in a distributed system configuration because we assume that one CNN partition is mapped onto one edge device.    


\subsection{Large-scale Investigation Results}
\label{sec:investigation_results}
\begin{figure}[!t]
	\includegraphics[width=\linewidth]{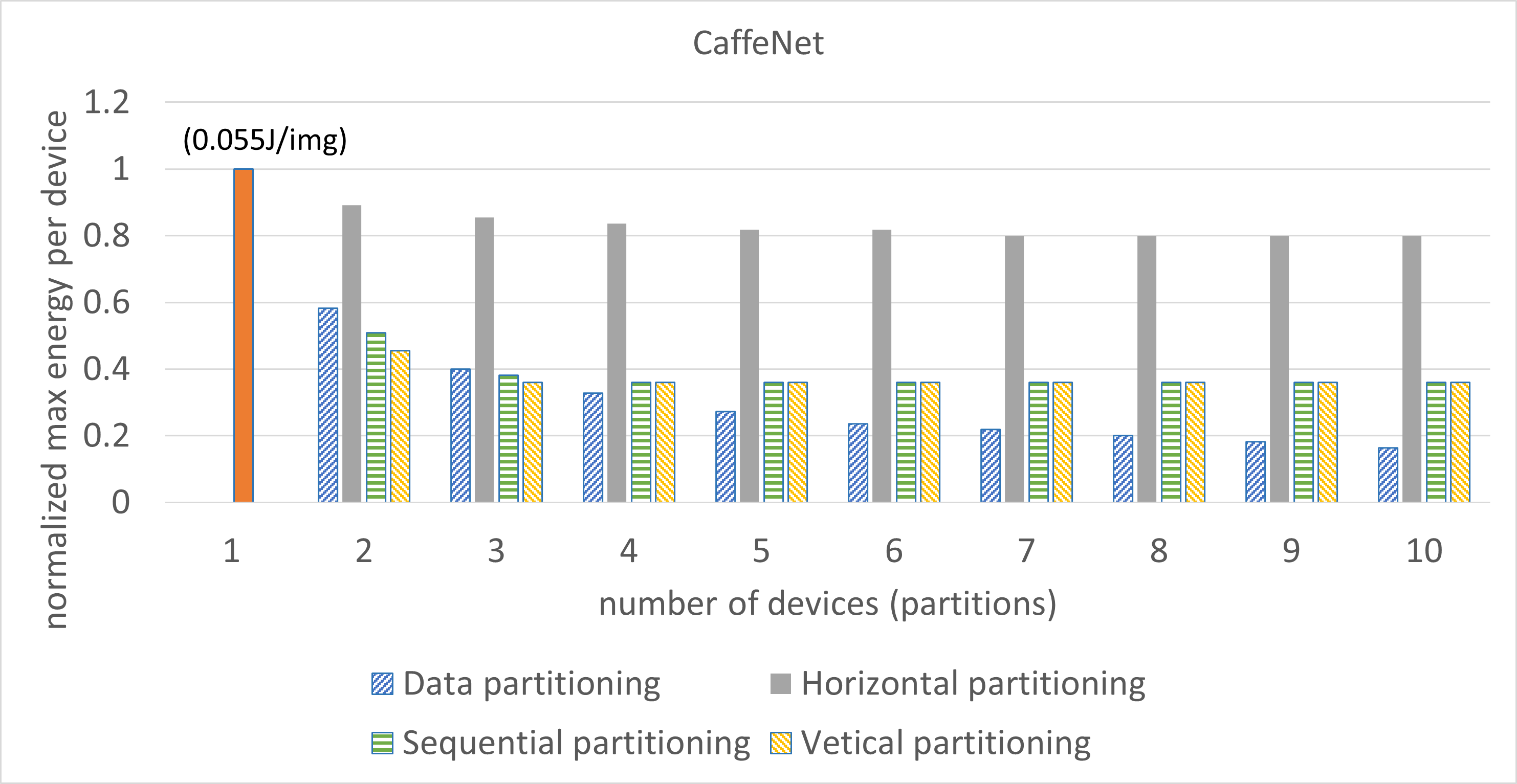}
	\caption{Per-device energy consumption for CaffeNet}
	\label{fig:energy_investigation_caffenet}
\end{figure}

\begin{figure}[!t]
	\includegraphics[width=\linewidth]{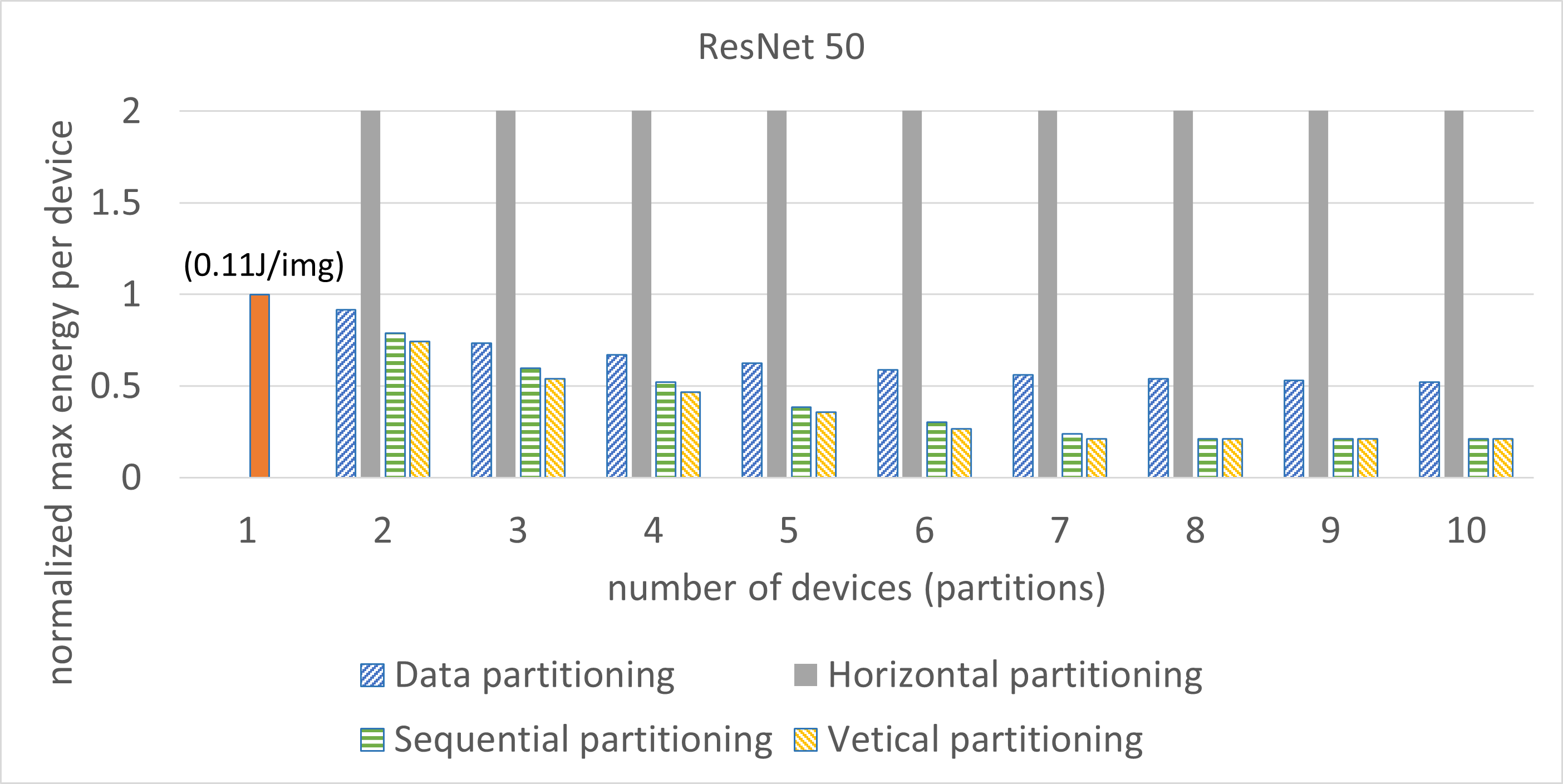}
	\caption{Per-device energy consumption for ResNet 50}
	\label{fig:energy_investigation_resnet50}
\end{figure}

\begin{figure}[!t]
	\includegraphics[width=\linewidth]{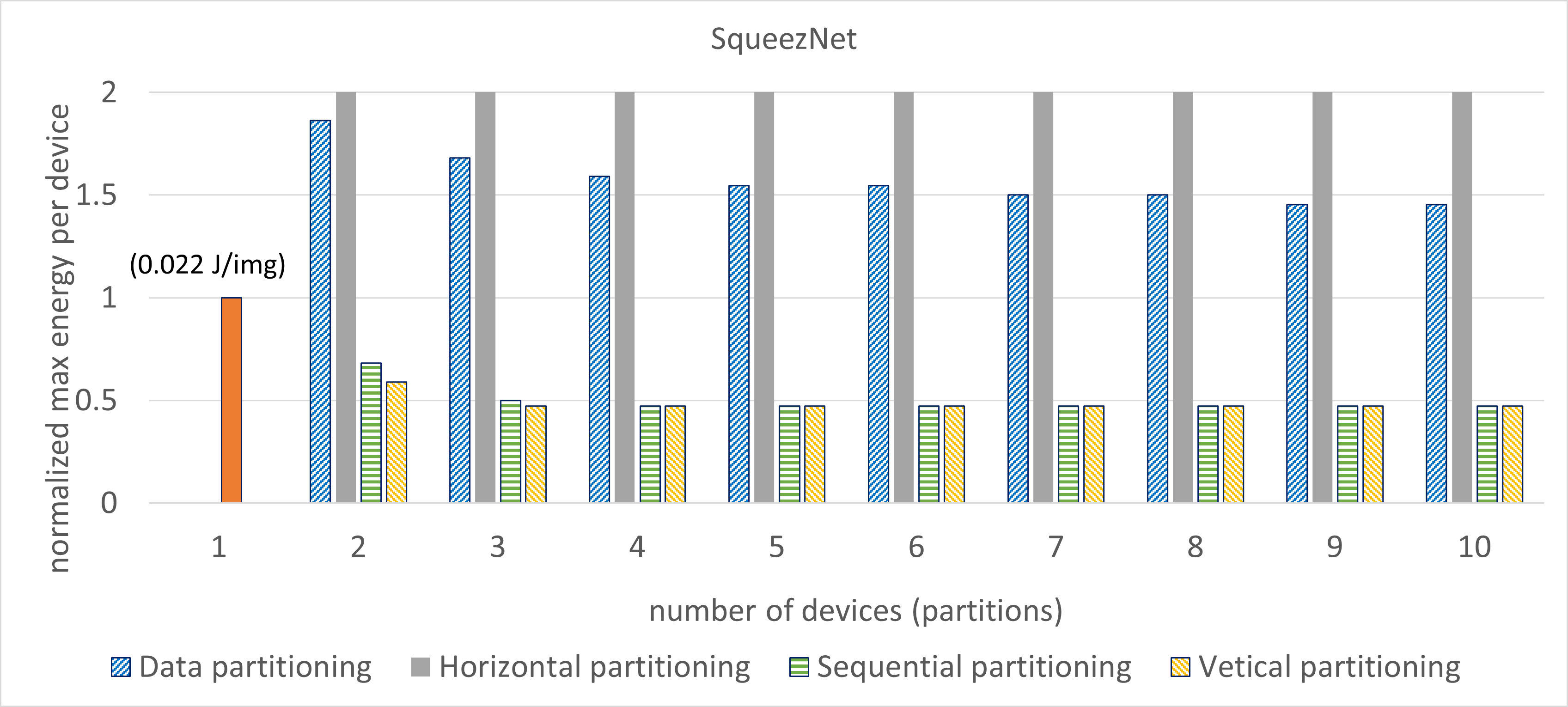}
	\caption{Per-device energy consumption for SqueezeNet}
	\label{fig:energy_investigation_squeeznet}
\end{figure}

\begin{figure}[!t]
	\includegraphics[width=\linewidth]{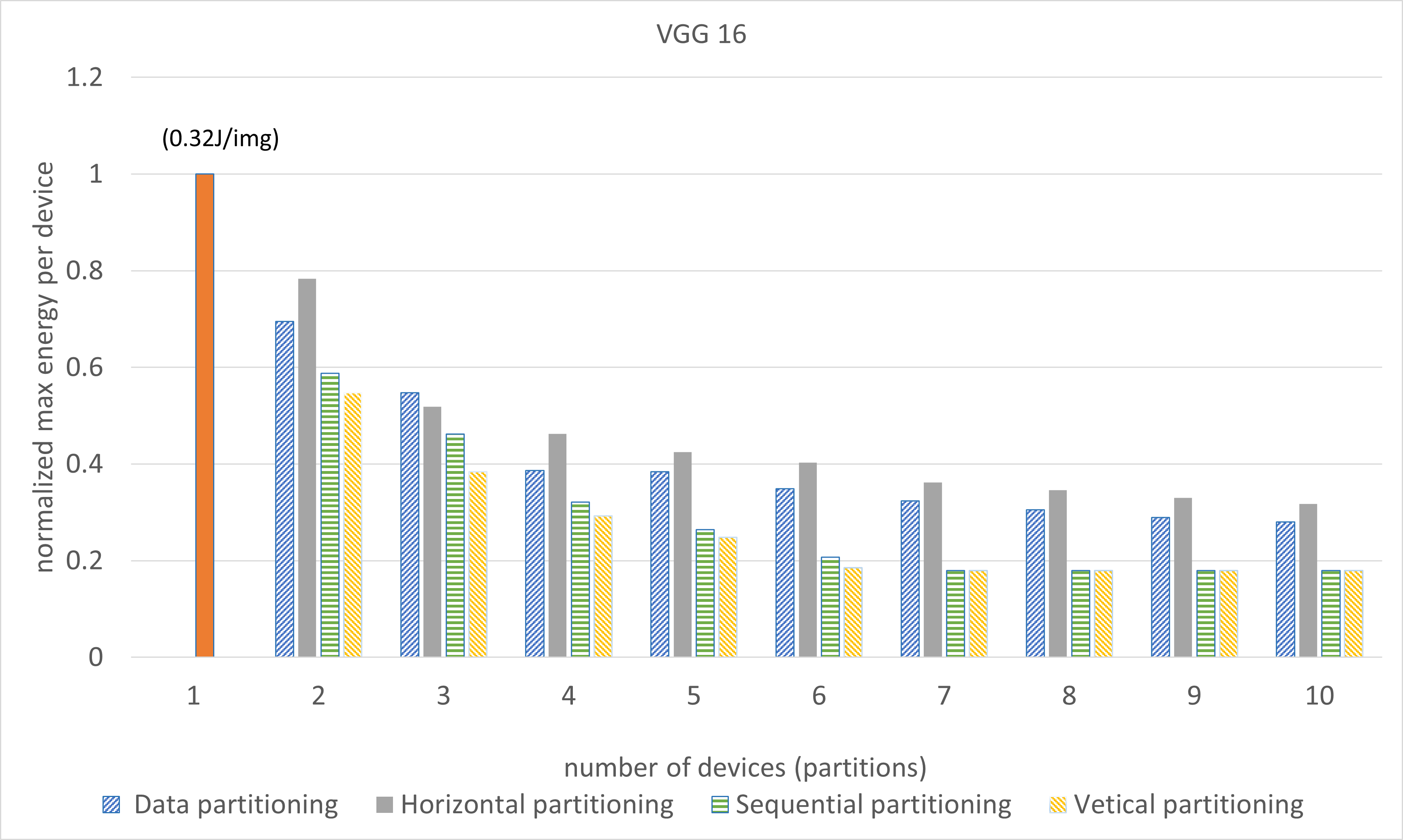}
	\caption{Per-device energy consumption for VGG 16}
	\label{fig:energy_investigation_vgg16}
\end{figure}

\begin{figure}[!t]
	\includegraphics[width=\linewidth]{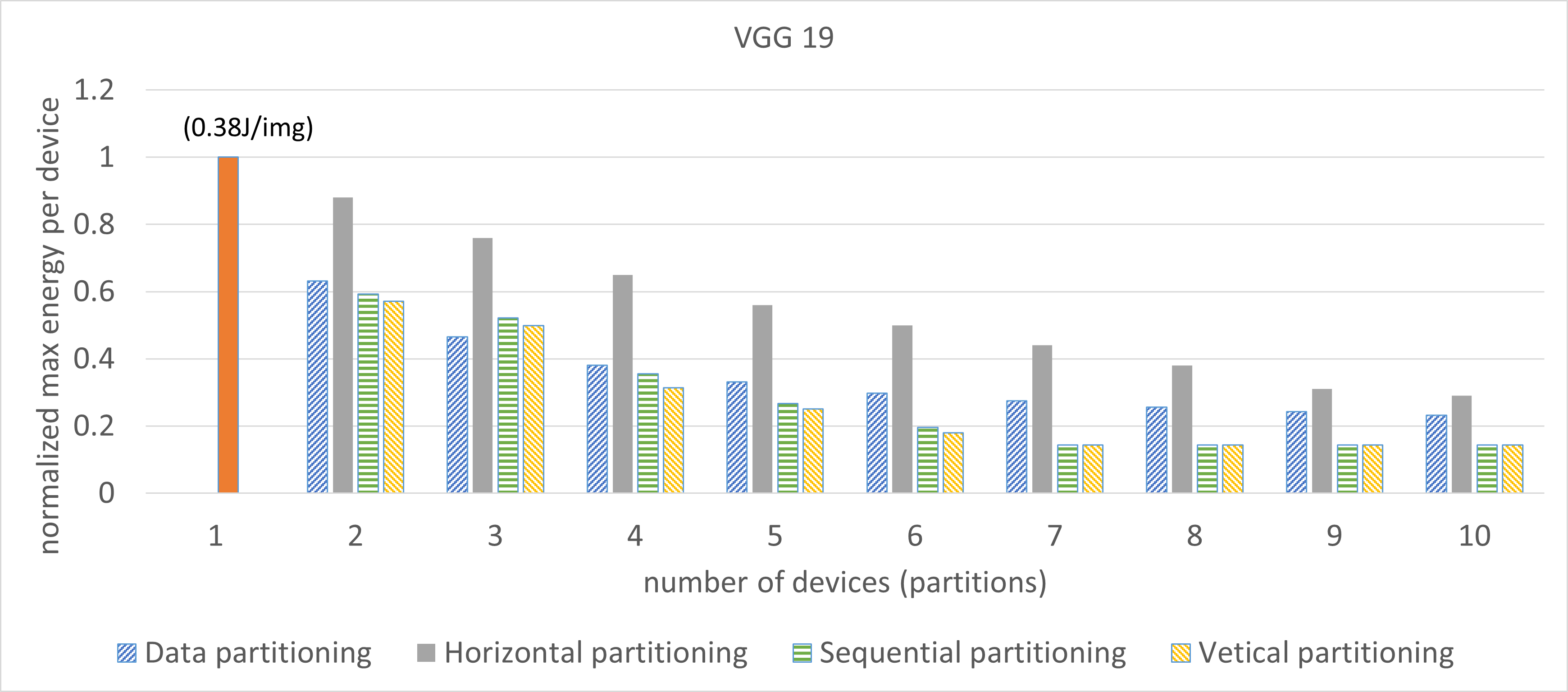}
	\caption{Per-device energy consumption for VGG 19}
	\label{fig:energy_investigation_vgg19}
\end{figure}

\begin{figure}[!t]
	\includegraphics[width=\linewidth]{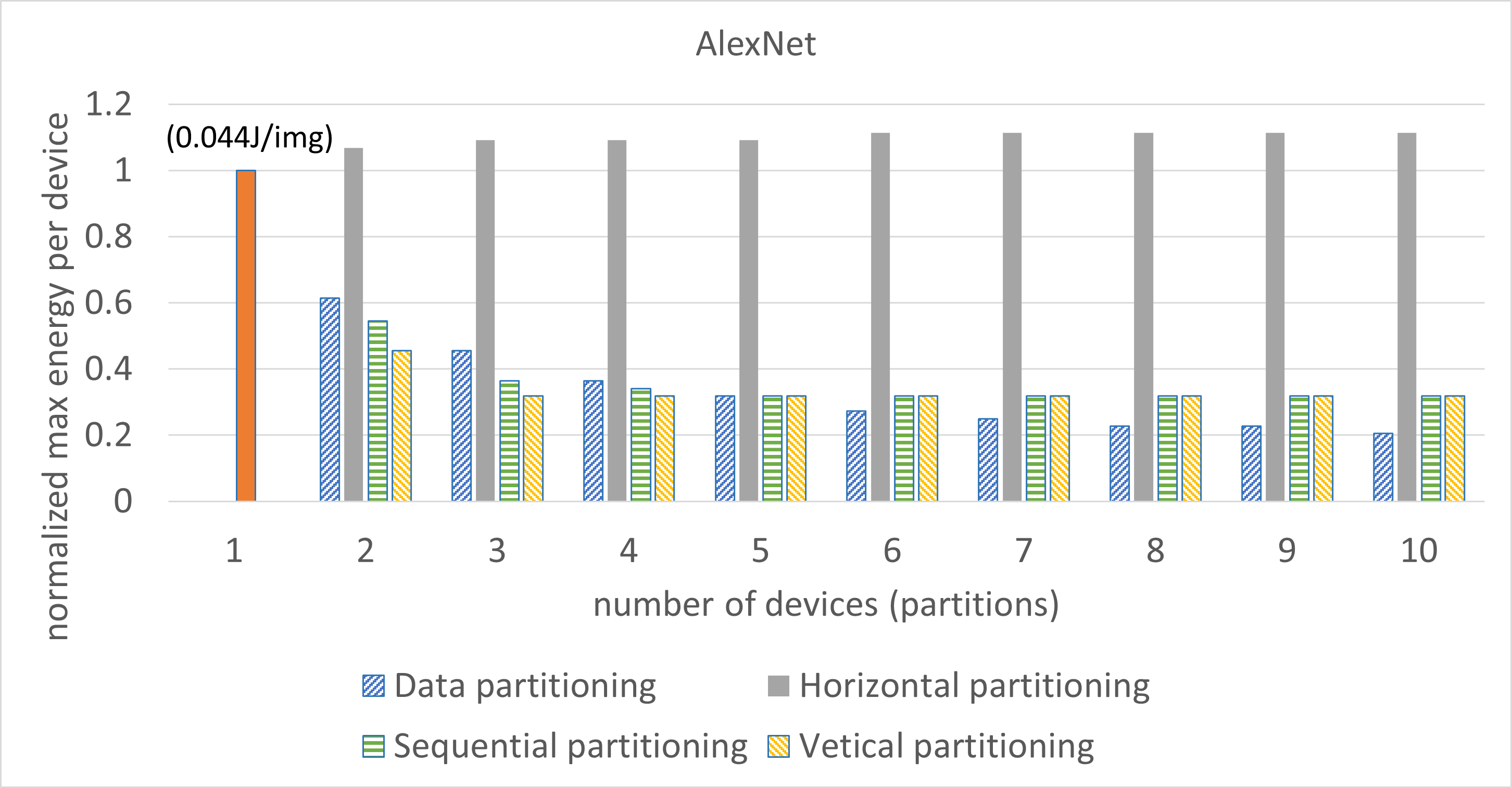}
	\caption{Per-device energy consumption for AlexNet}
	\label{fig:energy_investigation_alexnet}
\end{figure}

\begin{figure}[!t]
	\includegraphics[width=\linewidth]{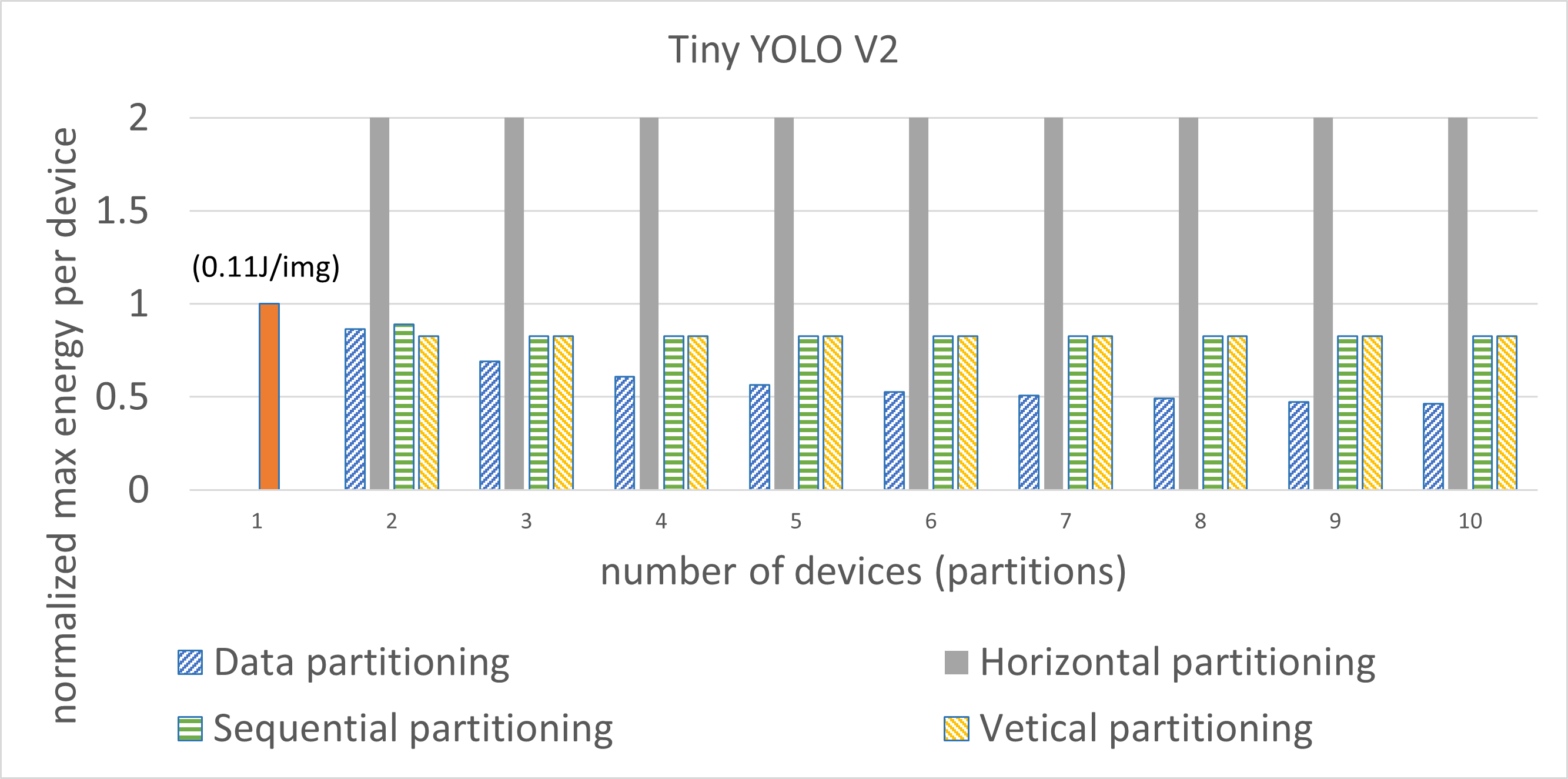}
	\caption{Per-device energy consumption for Tiny YOLO V2}
	\label{fig:energy_investigation_tiny_yolo}
\end{figure}

\begin{figure}[!t]
	\includegraphics[width=\linewidth]{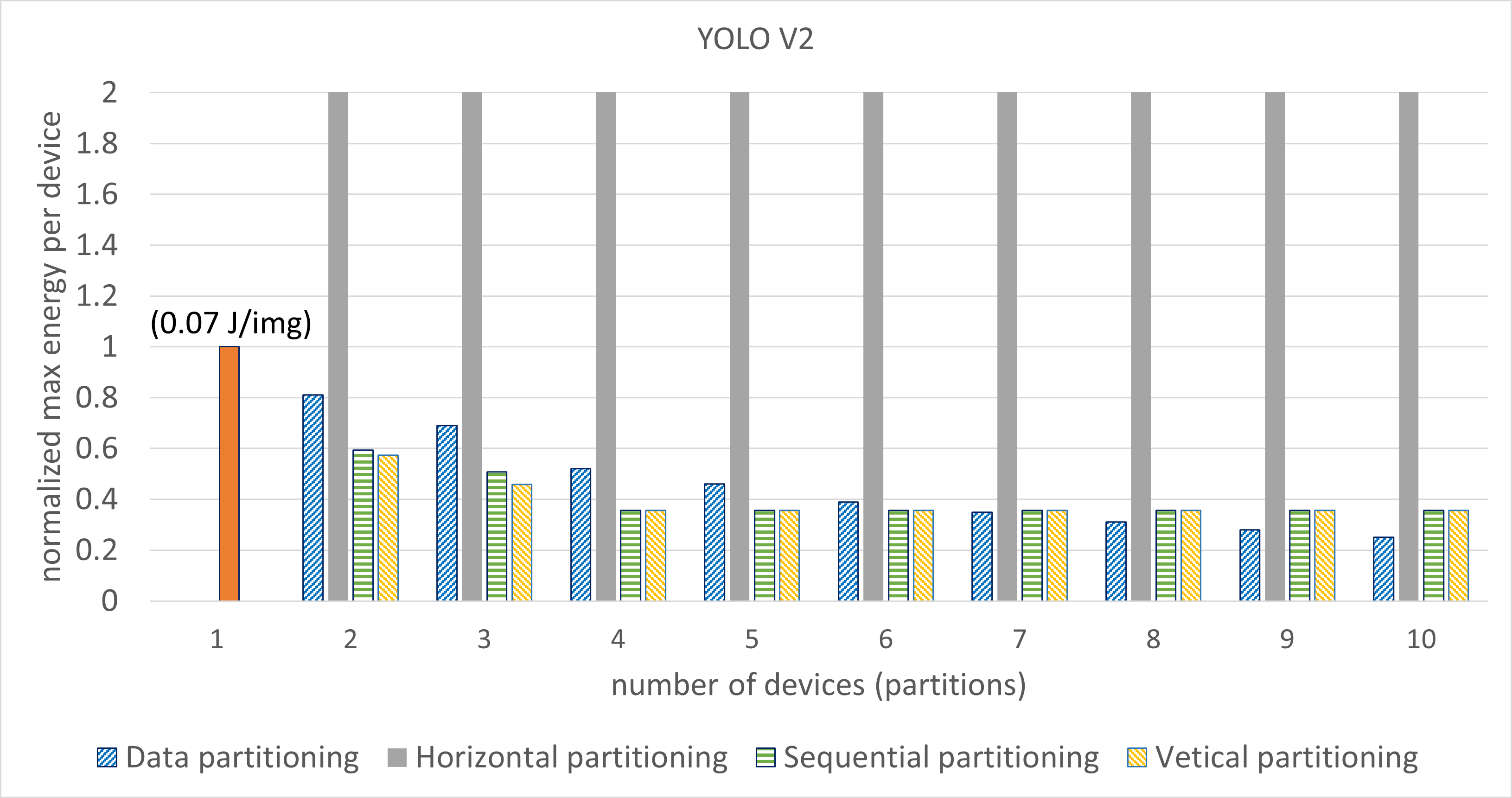}
	\caption{Per-device energy consumption for YOLO V2}
	\label{fig:energy_investigation_yolo}
\end{figure}

\begin{figure}[!t]
	\includegraphics[width=\linewidth]{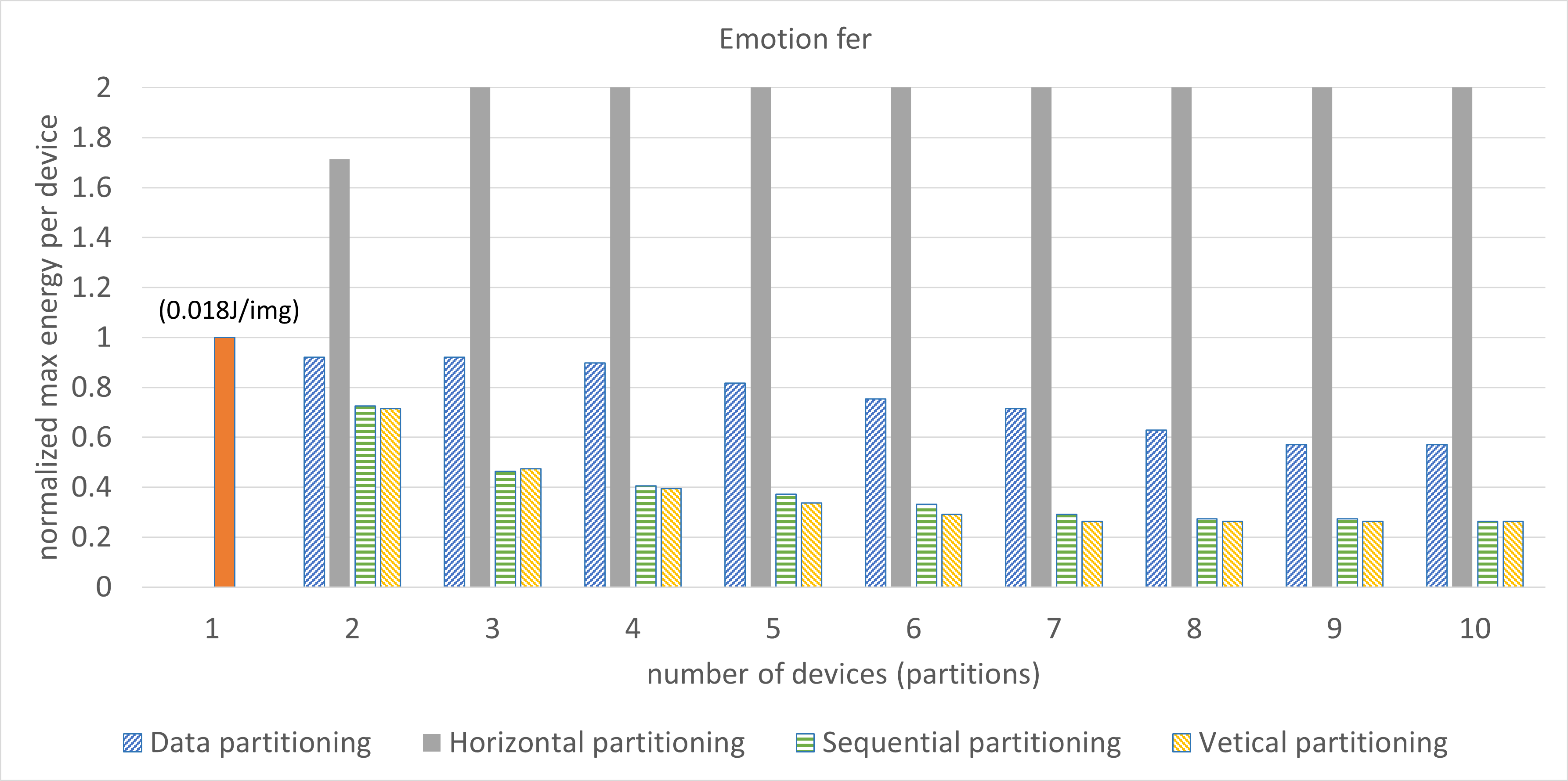}
	\caption{Per-device energy consumption for Emotion\_fer}
	\label{fig:energy_investigation_emotion_fer}
\end{figure}

In this section, we present the experimental results obtained during our investigation. For the nine representative CNN models, the results are plotted in nine separate charts shown in Figure~\ref{fig:energy_investigation_caffenet} to Figure~\ref{fig:energy_investigation_emotion_fer}. The horizontal axis in every chart shows the number of edge devices/partitions considered in a distributed system configuration to perform distributed CNN model inference. We consider distributed system configurations with 2 to 10 edge devices where, for every distributed system configuration, we plot four bars corresponding to the four partitioning strategies considered in our investigation. Every bar shows the maximum energy consumption per edge device of a CNN model inferred on a distributed system configuration, normalized to the energy consumption of the same CNN model inferred on one edge device.

Figure~\ref{fig:energy_investigation_caffenet} shows how the maximum energy consumption per edge device is affected by scaling up the number of edge devices in a distributed system configuration when the CaffeNet model is distributed by utilizing the four partitioning strategies. More specifically, the series of blue, grey, green, and light orange bars show the effects on the per-device energy consumption when utilizing the data partitioning strategy, horizontal partitioning strategy, sequential partitioning strategy, and vertical partitioning strategy, respectively. From Figure~\ref{fig:energy_investigation_caffenet}, we can see that for CaffeNet the horizontal partitioning strategy does not help to significantly decrease the per-device energy consumption when the number of edge devices/partitions increases. On the other hand, the data partitioning strategy keeps the per-device energy consumption decreasing with the increase of the number of edge devices and the decrease of per-device energy consumption can be significant, e.g., the decrease can reach 80\% in a distributed system with 8 edge devices. The sequential and vertical partitioning strategies perform slightly better than the data partitioning strategy in decreasing the per-device energy consumption when the number of edge devices/partitions is up to 3.  However, since these partitioning strategies cannot partition a single CNN layer, and CaffeNet has at least one large layer which is individually mapped on one edge device and determines the maximum per-device energy consumption, then increasing the number of edge devices/partitions beyond 3 does not help decreasing the maximum per-device energy consumption further.

Figure~\ref{fig:energy_investigation_resnet50} shows the experimental result for the ResNet 50 model. Here, the horizontal partitioning strategy performs very bad, i.e., the per-device energy consumption increases when the number of edge devices/partitions scales up.  Indeed, when the numbers of devices/partitions in a distributed system configuration is 2 or higher, the energy consumption per edge device actually increases more than two times compared to a system with only one edge device. This is indicated by the grey bars in Figure~\ref{fig:energy_investigation_resnet50} that are cut to value 2 in order to keep the chart more readable. Comparing the partitioning strategies, Figure~\ref{fig:energy_investigation_resnet50} shows that for ResNet 50 the vertical partitioning strategy is the best in terms of decreasing the energy consumption per edge device, with a maximum decrease of 80\% achieved on distributed systems containing more than 8 edge devices.

Figure~\ref{fig:energy_investigation_squeeznet} shows the experimental result for the SqueezeNet model. We can see that both the horizontal partitioning strategy and the data partitioning strategy increase the energy consumption per edge device when we infer SqueezeNet on a distributed system configuration with 2 or more edge devices. This is because an edge device/partition has to communicate a lot of data with other edge devices/partitions in the distributed system, thereby consuming a lot of additional energy. In addition, Figure~\ref{fig:energy_investigation_squeeznet} shows that the sequential and vertical partitioning strategies perform equally good and can effectively decrease the per-device energy consumption of the SqueezeNet model inferred on a distributed system configuration with up to 4 edge devices/partitions.   

Figure~\ref{fig:energy_investigation_vgg16} and Figure~\ref{fig:energy_investigation_vgg19} show our experimental result for the VGG 16 and VGG 19 models, respectively. For both models, the results indicate that the data, sequential, and vertical partitioning strategies could help to significantly decrease the energy consumption per edge device when the numbers of devices/partitions increases. Among these three strategies, the vertical partitioning strategy shows the best trend for per-device energy reduction. Considering the results for VGG 16 and VGG 19 with respect to the horizontal partitioning strategy, we conclude that this strategy can somewhat decrease the per-device energy consumption but it is less effective compared to the other three strategies.

Figure~\ref{fig:energy_investigation_alexnet} shows the experimental result for AlexNet. We see that similar to CaffeNet, when the numbers of devices in a distributed system is less than 5 (for CaffeNet less than 4), the vertical partitioning strategy is the best in terms of per-device energy reduction. When the numbers of devices/partitions is 5 or higher, the data partitioning strategy becomes the best because AlexNet has large layers with large input/output data tensors that cannot be partitioned by the vertical partitioning strategy to further decrease the per-device energy consumption. 

Figure~\ref{fig:energy_investigation_tiny_yolo} shows the experimental results for the Tiny YOLO V2 model. When the numbers of devices/partitions is 2 then the data, sequential, and vertical partitioning strategies achieve  similar per-device energy reduction of about 20\%. By further increasing the numbers of devices/partitions, we observe that the data partitioning strategy continues to decrease the maximum energy consumption per device whereas the sequential and vertical partitioning strategies do not further reduce the per-device energy consumption. This is because the data partitioning strategy is able to continuously  partition the input/output data tensors of layers thereby reducing the amount of computations performed by the layers in a partition and decreasing the energy consumption per device/partition.  

Figure~\ref{fig:energy_investigation_yolo} shows the experimental results for YOLO V2. Similar to AlexNet and CaffeNet, we observe that when the numbers of devices/partitions is less than a certain number, in this case 7, the vertical partitioning strategy is the best in terms of decreasing the energy consumption per edge device with a maximum decrease of about 65\%. By further increasing the numbers of devices/partitions, we see that the data partitioning strategy continues to slightly decrease the maximum energy consumption per device whereas the sequential and vertical partitioning strategies do not further reduce the per-device energy consumption. The reason for this trend is the same as explained above for Tiny YOLO V2.

Finally, Figure~\ref{fig:energy_investigation_emotion_fer} shows the experimental result for the Emotion\_fer model.  We observe that the vertical partitioning strategy achieves the best per-device energy reduction in all cases, with a maximum decrease of about 75\%, although the reduction stops when the numbers of devices/partitions is more than 7.

In Figure~\ref{fig:energy_investigation_caffenet} to Figure~\ref{fig:energy_investigation_emotion_fer}, the number in brackets above the red bar shows the energy consumption in Joules per image (J/img) when the CNN is executed stand-alone on one edge device. In our experiments, we use Jetson TX2 platforms/boards as edge devices and we power each Jetson TX2 board by a battery with capacity 18000 mAh and output voltage 19 V.  With this battery capacity, a stand-alone execution on one edge device of the smallest CNN (Emotion\_fer in Figure~\ref{fig:energy_investigation_emotion_fer}) can last only for 4.6 days, thereby processing 70354286 images. This is because the execution time for processing one image is 5.65 ms and the energy consumption per image is 0.0175 J. The execution of the largest CNN (VGG 19 in Figure~\ref{fig:energy_investigation_vgg19}) on one edge device can last only for 3.2 days, thereby processing 3214621 images. This is because the execution time for processing one image is 85.2 ms and the energy consumption per image is 0.383 J. 

Let us assume an application scenario with a requirement that the CNN execution must last for at least 7 days on a single battery charge. If the aforementioned CNNs (Emotion\_fer and VGG 19) are executed stand-alone on one Jetson TX2 edge device powered by the aforementioned battery then the 7-day requirement will not be met. So, our assumed scenario and requirement is a clear motivation that we have to partition the CNNs and execute each CNN in a distributed manner on multiple edge devices in order to meet the requirement. According to our investigation and results for Emotion\_fer in Figure~\ref{fig:energy_investigation_emotion_fer}, we have to partition Emotion\_fer on more than 3 edge devices in order to reduce the energy consumption per device with at least 35\% which is sufficient to meet our 7-day requirement. In addition, we can see that the vertical partitioning strategy is the best choice to be utilized because it saves the most energy consumption. According to our investigation and results for VGG 19 in Figure~\ref{fig:energy_investigation_vgg19}, we have to partition VGG 19 on more than 4 edge devices, thereby achieving at least 55\% energy reduction per device, which is sufficient to meet our requirement. Again, the vertical partitioning strategy is the best choice to be utilized.

\subsection{Discussion}
\label{sec:Discussion}
Based on the large-scale investigation results, presented in Section~\ref{sec:investigation_results}, we could state the following general findings on how the four partitioning strategies perform in terms of decreasing the maximum energy consumption per edge device, when large CNN models are inferred on a distributed system at the edge:
\begin{itemize}
\item The Horizontal Partitioning Strategy shows the worst performance among the four partitioning strategies, thus it should not be used as a tool for energy reduction in distributed CNN inference at the edge. Moreover, for most of the representative CNN models, this strategy actually increases the energy consumption per edge device when the number of devices/partitions is increased. This is because the Horizontal Partitioning Strategy requires every CNN layer, belonging to a partition (edge device), to send its output data to all other partitions/devices. Such external data communication significantly increases the energy consumption of an edge device in a distributed system.
\item The Vertical Partitioning Strategy shows equal or better performance compared to the Sequential Partitioning Strategy in terms of reduced energy consumption per edge device. This is because the Vertical Partitioning Strategy provides more flexibility in terms of possible partitioning configurations due to the fact that each partition in this strategy may include consecutive and  non-consecutive CNN layers whereas each partition in the Sequential Partitioning Strategy must include only consecutive CNN layers. Therefore, the Vertical Partitioning Strategy should always be preferred to the Sequential Partitioning Strategy. 
\item Comparing the Vertical Partitioning Strategy with the Data Partitioning Strategy, there is not a clear winner in terms of energy reduction potential. For many of the CNN models, the Vertical Partitioning Strategy shows the highest potential to reduce the per-device energy consumption irrespective of the number of devices/partitions. However, for some CNN models (see Figure~\ref{fig:energy_investigation_caffenet},~\ref{fig:energy_investigation_alexnet},~\ref{fig:energy_investigation_tiny_yolo}, and~\ref{fig:energy_investigation_yolo}), when the devices/partitions go beyond a certain number, the Data Partitioning Strategy leads to higher per-device energy reduction because this strategy is able to further reduce the computation and communication workload per device whereas the Vertical Partitioning Strategy cannot achieve this.
\end{itemize}
\section{Conclusions}
\label{sec:Conclusions}

In this paper, we have investigated, compared, and discussed the energy consumption per edge device when a CNN model is partitioned and inferred on a distributed system at the edge by applying the existing four partitioning strategies, namely Data Partitioning Strategy, Horizontal Partitioning Strategy, Sequential Partitioning Strategy, and Vertical Partitioning Strategy. In order to facilitate our investigation, we have proposed novel and very accurate analytical models to estimate the energy consumption of every CNN partition when a CNN model is partitioned by utilizing the four partitioning strategies. The proposed energy consumption analytical models have been validated by comparing the energy consumption of partitioned CNNs, estimated by these analytical models, with the corresponding energy consumption numbers, obtained by direct measurements when the same partitioned CNNs are deployed and executed on real distributed system configurations. 

Our accurate energy consumption analytical models have allowed us to conduct a large-scale experiment including nine representative real-world CNN models, taken from the ONNX model zoo library, to investigate which partitioning strategy has the highest potential to decrease the per-device energy consumption for every one of these nine CNN models inferred on distributed devices at the edge. From the investigation and the obtained experimental results we conclude that: 1) Horizontal Partitioning Strategy should not be used to reduce the per-device energy consumption in distributed CNN inference at the edge because this strategy has shown the lowest potential among the four partitioning strategies; 2) In most cases, Vertical Partitioning Strategy has shown the highest potential among the four partitioning strategies, thus it can always be used with great confidence to reduce the per-device energy consumption; 3) Only in specific cases, Data Partitioning Strategy has shown better results than Vertical Partitioning Strategy in reducing the per-device energy consumption, i.e., in specific cases where the size of input/output tensors of CNN layers is relatively large and the number of partitions/devices is also large.


%
\balance
\bibliography{IEEEabrv,./main.bib}

\begin{thebibliography}{10}
\providecommand{\url}[1]{#1}
\csname url@samestyle\endcsname
\providecommand{\newblock}{\relax}
\providecommand{\bibinfo}[2]{#2}
\providecommand{\BIBentrySTDinterwordspacing}{\spaceskip=0pt\relax}
\providecommand{\BIBentryALTinterwordstretchfactor}{4}
\providecommand{\BIBentryALTinterwordspacing}{\spaceskip=\fontdimen2\font plus
\BIBentryALTinterwordstretchfactor\fontdimen3\font minus
  \fontdimen4\font\relax}
\providecommand{\BIBforeignlanguage}[2]{{%
\expandafter\ifx\csname l@#1\endcsname\relax
\typeout{** WARNING: IEEEtran.bst: No hyphenation pattern has been}%
\typeout{** loaded for the language `#1'. Using the pattern for}%
\typeout{** the default language instead.}%
\else
\language=\csname l@#1\endcsname
\fi
#2}}
\providecommand{\BIBdecl}{\relax}
\BIBdecl

\bibitem{b1}
M.~Z. Alom, T.~M. Taha, C.~Yakopcic, S.~Westberg, P.~Sidike, M.~S. Nasrin,
  B.~C. Van~Esesn, A.~A.~S. Awwal, and V.~K. Asari, ``The history began from
  alexnet: A comprehensive survey on deep learning approaches,'' \emph{arXiv
  preprint arXiv:1803.01164}, 2018.

\bibitem{b2}
Y.~Cheng, D.~Wang, P.~Zhou, and T.~Zhang, ``A survey of model compression and
  acceleration for deep neural networks,'' \emph{arXiv preprint
  arXiv:1710.09282}, 2017.

\bibitem{b3}
S.~Han, H.~Mao, and W.~J. Dally, ``Deep compression: Compressing deep neural
  networks with pruning, trained quantization and huffman coding,'' \emph{arXiv
  preprint arXiv:1510.00149}, 2015.

\bibitem{b4}
S.~Yao, Y.~Zhao, A.~Zhang, L.~Su, and T.~Abdelzaher, ``Deepiot: Compressing
  deep neural network structures for sensing systems with a compressor-critic
  framework,'' in \emph{Proceedings of the 15th ACM Conference on Embedded
  Network Sensor Systems}, 2017, pp. 1--14.

\bibitem{b5}
N.~Kalatzis, M.~Avgeris, D.~Dechouniotis, K.~Papadakis-Vlachopapadopoulos,
  I.~Roussaki, and S.~Papavassiliou, ``Edge computing in iot ecosystems for
  uav-enabled early fire detection,'' in \emph{2018 IEEE international
  conference on smart computing (SMARTCOMP)}.\hskip 1em plus 0.5em minus
  0.4em\relax IEEE, 2018, pp. 106--114.

\bibitem{b6}
Y.~Kang, J.~Hauswald, C.~Gao, A.~Rovinski, T.~Mudge, J.~Mars, and L.~Tang,
  ``Neurosurgeon: Collaborative intelligence between the cloud and mobile
  edge,'' \emph{ACM SIGARCH Computer Architecture News}, vol.~45, no.~1, pp.
  615--629, 2017.

\bibitem{b7}
Z.~Zhao, K.~M. Barijough, and A.~Gerstlauer, ``Deepthings: Distributed adaptive
  deep learning inference on resource-constrained iot edge clusters,''
  \emph{IEEE Transactions on Computer-Aided Design of Integrated Circuits and
  Systems}, vol.~37, no.~11, pp. 2348--2359, 2018.

\bibitem{b8}
S.~Wang, G.~Ananthanarayanan, Y.~Zeng, N.~Goel, A.~Pathania, and T.~Mitra,
  ``High-throughput cnn inference on embedded arm big. little multicore
  processors,'' \emph{IEEE Transactions on Computer-Aided Design of Integrated
  Circuits and Systems}, vol.~39, no.~10, pp. 2254--2267, 2019.

\bibitem{b9}
M.~Jouhari, A.~K. Al-Ali, E.~Baccour, A.~Mohamed, A.~Erbad, M.~Guizani, and
  M.~Hamdi, ``Distributed cnn inference on resource-constrained uavs for
  surveillance systems: Design and optimization,'' \emph{IEEE Internet of
  Things Journal}, vol.~9, no.~2, pp. 1227--1242, 2021.

\bibitem{b10}
R.~Stahl, Z.~Zhao, D.~Mueller-Gritschneder, A.~Gerstlauer, and U.~Schlichtmann,
  ``Fully distributed deep learning inference on resource-constrained edge
  devices,'' in \emph{International Conference on Embedded Computer
  Systems}.\hskip 1em plus 0.5em minus 0.4em\relax Springer, 2019, pp. 77--90.

\bibitem{b11}
J.~Mao, X.~Chen, K.~W. Nixon, C.~Krieger, and Y.~Chen, ``Modnn: Local
  distributed mobile computing system for deep neural network,'' in
  \emph{Design, Automation \& Test in Europe Conference \& Exhibition (DATE),
  2017}.\hskip 1em plus 0.5em minus 0.4em\relax IEEE, 2017, pp. 1396--1401.

\bibitem{b12}
S.~Bhattacharya and N.~D. Lane, ``Sparsification and separation of deep
  learning layers for constrained resource inference on wearables,'' in
  \emph{Proceedings of the 14th ACM Conference on Embedded Network Sensor
  Systems CD-ROM}, 2016, pp. 176--189.

\bibitem{b19}
E.~Tang and T.~Stefanov, ``Low-memory and high-performance cnn inference on
  distributed systems at the edge,'' in \emph{Proceedings of the 14th IEEE/ACM
  International Conference on Utility and Cloud Computing Companion}, 2021, pp.
  1--8.

\bibitem{b14}
\BIBentryALTinterwordspacing
ONNX. (2022) Onnx model zoo. [Online]. Available:
  \url{https://github.com/onnx/models}
\BIBentrySTDinterwordspacing

\bibitem{b15}
S.~Zhang, S.~Zhang, Z.~Qian, J.~Wu, Y.~Jin, and S.~Lu, ``Deepslicing:
  collaborative and adaptive cnn inference with low latency,'' \emph{IEEE
  Transactions on Parallel and Distributed Systems}, vol.~32, no.~9, pp.
  2175--2187, 2021.

\bibitem{b17}
K.~Vanishree, A.~George, S.~Gunisetty, S.~Subramanian, S.~Kashyap, and
  M.~Purnaprajna, ``Coin: Accelerated cnn co-inference through data
  partitioning on heterogeneous devices,'' in \emph{2020 6th International
  Conference on Advanced Computing and Communication Systems (ICACCS)}.\hskip
  1em plus 0.5em minus 0.4em\relax IEEE, 2020, pp. 90--95.

\bibitem{b13}
E.~Tang, S.~Minakova, and T.~Stefanov, ``Energy-efficient and high-throughput
  cnn inference on embedded cpus-gpus mpsocs,'' in \emph{International
  Conference on Embedded Computer Systems}.\hskip 1em plus 0.5em minus
  0.4em\relax Springer, 2022, pp. 127--143.

\bibitem{b20}
J.~Mas, T.~Panadero, G.~Botella, A.~A. Del~Barrio, and C.~Garc{\'\i}a, ``Cnn
  inference acceleration using low-power devices for human monitoring and
  security scenarios,'' \emph{Computers \& Electrical Engineering}, vol.~88, p.
  106859, 2020.

\bibitem{b21}
S.~Y.~H. Mirmahaleh, M.~Reshadi, H.~Shabani, X.~Guo, and N.~Bagherzadeh, ``Flow
  mapping and data distribution on mesh-based deep learning accelerator,'' in
  \emph{Proceedings of the 13th IEEE/ACM International Symposium on
  Networks-on-Chip}, 2019, pp. 1--8.

\bibitem{b22}
R.~Xie, X.~Jia, L.~Wang, and K.~Wu, ``Energy efficiency enhancement for
  cnn-based deep mobile sensing,'' \emph{IEEE Wireless Communications},
  vol.~26, no.~3, pp. 161--167, 2019.

\bibitem{b23}
M.~Abadi, M.~Isard, and D.~G. Murray, ``A computational model for tensorflow:
  an introduction,'' in \emph{Proceedings of the 1st ACM SIGPLAN International
  Workshop on Machine Learning and Programming Languages}, 2017, pp. 1--7.

\bibitem{b26}
F.~Darema, D.~A. George, V.~A. Norton, and G.~F. Pfister, ``A
  single-program-multiple-data computational model for epex/fortran,''
  \emph{Parallel Computing}, vol.~7, no.~1, pp. 11--24, 1988.

\bibitem{b27}
P.~Pirsch and H.~Jeschke, ``Mimd (multiple instruction multiple data)
  multiprocessor system for real-time image processing,'' in \emph{Image
  Processing Algorithms and Techniques II}, vol. 1452.\hskip 1em plus 0.5em
  minus 0.4em\relax SPIE, 1991, pp. 544--555.

\bibitem{b28}
``Nvidia jetson tx2,'' \emph{https://developer.nvidia.com/embedded/jetson-tx2}.

\bibitem{b29}
L.~Clarke, I.~Glendinning, and Hempel, ``The mpi message passing interface
  standard,'' \emph{Programming environments for massively parallel distributed
  systems}, vol.~7, no.~1, pp. 213--218, 1994.

\end{thebibliography}

\end{document}